\documentclass[nohyperref]{article}

\usepackage{microtype}
\usepackage{graphicx}
\usepackage{natbib}
\usepackage{caption}
\usepackage[dvipsnames]{xcolor}
\usepackage{subfigure}
\usepackage{booktabs} % for professional tables
\usepackage{multicol}
\usepackage{multirow}
\usepackage{tabularx}

\newcommand\corr[1]{\textcolor{black}{#1}}
\newcommand\corrr[1]{\textcolor{black}{#1}}
\newcommand\renata[1]{\textcolor{black}{#1}}
\newcommand\kenyu[1]{\textcolor{black}{#1}}
\newcommand\arno[1]{\textcolor{black}{#1}}
\newcommand\eq[1]{Eq.~(\ref{#1})}
\newcommand\sect[1]{Sec.~\ref{#1}}

\usepackage{hyperref}

\usepackage[accepted]{icml2023}

\usepackage{amsmath}
\usepackage{amssymb}
\usepackage{mathtools}
\usepackage{amsthm}

\usepackage[capitalize,noabbrev]{cleveref}

\theoremstyle{plain}

\theoremstyle{definition}

\theoremstyle{remark}

\icmltitlerunning{Unlocking LRP for Autoencoders}

\begin{document}

\twocolumn[
% \icmltitle{Reconstruction Error-based Explainability for Autoencoder models}
% \icmltitle{Efficient Explainability solution for Autoencoders}
\icmltitle{Unlocking Layer-wise Relevance Propagation for Autoencoders}

%\icmltitle{LRP for Reconstruction Error in Autoencoders}
%\icmltitle{Extension of Layer-Wise Relevance Propagation to Explain Unsupervised Autoencoders}



\icmlsetsymbol{equal}{*}

\begin{icmlauthorlist}
\icmlauthor{Kenyu Kobayashi}{equal,labs}
\icmlauthor{Renata Khasanova}{equal,labs}
\icmlauthor{Arno Schneuwly}{equal,labs}
\icmlauthor{Felix Schmidt}{labs}
\icmlauthor{Matteo Casserini}{labs}
\end{icmlauthorlist}

\icmlaffiliation{labs}{Oracle Labs, Switzerland}

\icmlcorrespondingauthor{Kenyu Kobayashi}{kenyu.kobayashi@oracle.com}
\icmlkeywords{Machine Learning Explainability, Anomaly Detection, Autoencoders, Backpropagation, Layer-Wise-Relevance Propagation, Reconstruction Loss, Deep Taylor Decomposition}

\vskip 0.3in
]



\printAffiliationsAndNotice{\icmlEqualContribution} % otherwise use the standard text.
%A  gateway to find anomalies with unsupervised anomaly detection in autoencoders is the reconstruction error.
%we can introduce that antoencoder is used for multiple problems, however, it is not trivial to explain them. Therefore, we propose fast explainability solution by extending layer-wise propagation method and deep taylor decomposition framework. Our results highlight computational as well as qualitative advantages compared to existing methods.
\begin{abstract}
Autoencoders are a powerful and versatile tool often used for various problems such as anomaly detection, image processing and machine translation. However, their reconstructions are not always trivial to explain. Therefore, we propose a fast explainability solution by extending the Layer-wise Relevance Propagation method with the help of \corrr{the} Deep Taylor Decomposition framework. Furthermore, we introduce a novel validation technique for comparing our explainability approach with baseline methods in the case of missing ground-truth data. Our results highlight computational as well as qualitative advantages of the proposed explainability solution with respect to existing methods. 

\end{abstract}

\section{Introduction}
Autoencoders~\cite{autoencoders} are neural network architectures that play a fundamental role in unsupervised machine learning. They are frequently used in various tasks such as anomaly detection, machine translation and image processing~\cite{autoencoder_use_cases}. \corr{Autoencoders are designed to encode the input data into a compressed, meaningful representation. This representation is then decoded in such a way that the reconstruction is as close as possible to the input data.} \corrr{Specifically,} Autoencoders aim to minimize a reconstruction error, which is computed with a loss function based on the difference between the original input and its reconstruction. \corrr{By minimizing this error, Autoencoders learn the informative representation of the data.}
Nonetheless, despite their good performance and widespread usage across different applications, they are hardly interpretable due to their intrinsic nonlinearity.
%despite their good performance, they are hardly interpretable due to their nature. 
In particular, when an Autoencoder fails to properly reconstruct a given input, understanding the rationale behind this failure is challenging. To that end, the addition of explainability capabilities to these types of models is highly desirable.

\begin{figure}[!t]
    \begin{tabular}{cc}     
    \includegraphics[width=0.2\textwidth]{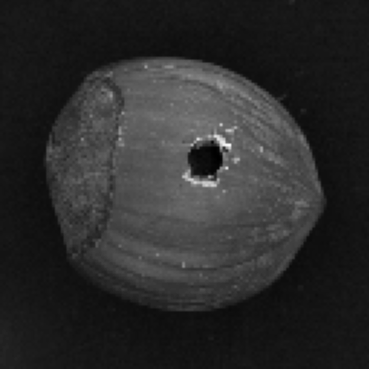}&
     \includegraphics[width=0.2\textwidth]{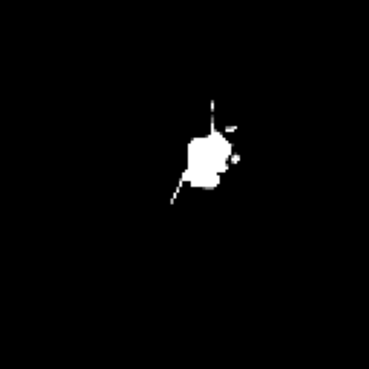}\\
     (a) Original & (b) Ground truth \\
    \includegraphics[width=0.2\textwidth]{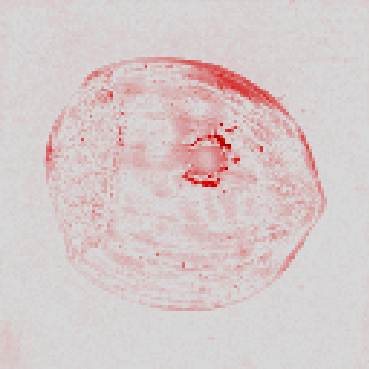}&
    \includegraphics[width=0.2\textwidth]{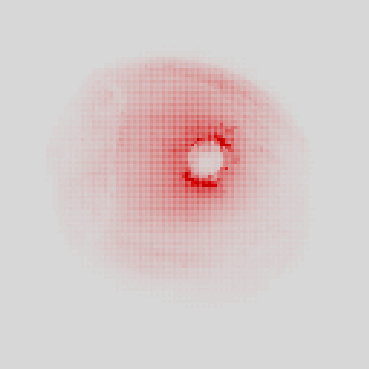}\\
    (c) Baseline & (d) Ours \\
    \end{tabular}
    \caption{Explanations produced for an image of a damaged object from the MVTec dataset~\cite{anomaly_image_dataset}. The figure illustrates the (a) original image, (b) ground truth damaged area, (c) explanations produced by the baseline method and (d) results of our LRP-based explainability approach. Here, the baseline method focuses not only on the damaged area of the hazelnut, but on the borders of the object as well. Our approach, on the other hand, focuses attention on the damaged area.} 
    \label{fig:LRPAE}
\end{figure}

One way to explain \arno{an} Autoencoder output for a given sample can be achieved \arno{by} using attribution-based explainability methods~\cite{xai_methods_survey}. The main idea of these approaches is to explain machine learning models by assigning a relevance score to each input feature depending on its importance for the model's prediction. Such scores can be computed by measuring the reconstruction error of each individual feature. But depending on the approach used, additional noise may appear in explanations, as depicted in Figure~\ref{fig:LRPAE} (c). This happens due to the fact that relevance scores are being assigned without taking into consideration any contextual information. Another approach is to modify the original input and measure the impact of such modifications on the model's output~\cite{shap, lime}. Then, high relevance scores are assigned to features that have a significant impact.

However, in this case the number of perturbations to reliably compute such scores increases exponentially with the dimensionality of \arno{the} input data. Hence, \arno{this process} may get computationally expensive. Other types of explainability approaches~\cite{lrp, deeplift, integratedgradients} leverage the knowledge of architecture configurations and often come with computational benefits. 
 
In this work we propose an explainability approach specific to Autoencoders by extending the Layer-wise Relevance Propagation (LRP) framework~\cite{lrp}. Figure~\ref{fig:LRPAE}~(d) illustrates an example of an explanation generated with our method when applied to a convolutional Autoencoder model. This model is trained to reconstruct images of a non-damaged sample object from the MVTec dataset~\cite{anomaly_image_dataset}.
 Figure~\ref{fig:LRPAE} shows that our explanation (d) is more focused on the damaged part, while the reconstruction error-based one~(c) focuses not only on the damaged area but also on borders and background noise.

Our contribution is two-fold:
\begin{itemize}
    \item \renata{we propose a novel LRP-based explainability approach specific to Autoencoders, that allows the propagation of reconstruction errors and assignment of relevance scores to input features};
	\item in order to assess explainability methods' performance, we introduce a self-supervised validation approach. The latter produces artificial explanation labels, which can be used for evaluation.
\end{itemize}

\section{Related Work}
\label{sec:related_work}
Attribution-based explanation methods~\cite{xai_methods_survey} have recently gained popularity in the field of explainable artificial intelligence~\cite{xai_dnn_survey}. The main idea of such approaches is to compute relevance scores for each input feature. The scores reflect the features corresponding importance with \arno{respect} to the model's decision-making process. Formally, given an input feature vector $x=\left[x_{1}, \ldots, x_{N}\right] \in \mathbb{R}^{N}$ and a model output $y$, the attribution-based approach assigns relevance scores $R = [R_1,...R_N]  \in \mathbb{R}^{N}$ to features, where $R_i$ designates the contribution of $x_i$ to the output $y$ of the model.

We distinguish between two classes of attribution-based methods: perturbation-based methods~\cite{shap, lime} and backpropagation-based methods~\cite{deeplift, integratedgradients, lrp}. While the former are model-agnostic and, therefore, applicable to any black-box model, the latter are model-specific and exploit the underlying structure of the model to provide explanations. Model-agnostic perturbation-based methods (e.g. SHAP~\cite{shap, adaeshap, adaeshap2}, LIME~\cite{lime}) typically analyze output values 
of multiple modifications of the original input feature vector, and assign relevance scores to features based on this analysis.
%of multiple modified versions of input feature vector and assign relevance scores to features. 
Therefore, they are computationally expensive~\cite{performance}, which makes them unsuitable for tasks \arno{with a high input dimensionality}. On the other hand, backpropagation-based approaches~\cite{backprop_based_methods} exploit the structure of predictive models to compute relevance scores in a single backward pass, which makes them computationally efficient. As these methods are model-specific, their field of application is constrained, as they require knowledge of the architecture.

\citet{deeplift} and \citet{integratedgradients} propose to estimate relevance scores based on differences between the model's output for a given sample and the model's output for what they call a baseline input. Such a baseline input needs to be selected \arno{with information on the} application domain. For example, \arno{in an image classification task, a black image may be selected as a baseline to represent the absence of any information.} Therefore, the main limitation of such an approach is the requirement of domain-specific knowledge. In contrast, \arno{the} LRP approach~\cite{lrp, lrpdifflayers} \arno{is} designed for neural network architectures and directly propagates the model's output backward using propagation rules designed for various layer types, such as fully-connected, pooling, convolutional layers, etc.~\cite{lrpexample1, lrpexample2, lrpexample3}.
The LRP explainability approach is \corrr{typically} applied to supervised tasks~\cite{lrpusecase1, lrpusecase6, lrpexample3, lrp_supervised_ae} and to, the best of our knowledge, no LRP propagation rule for the reconstruction loss function of Autoencoders exists.
\renata{Therefore, in this work, we propose such a rule that permits propagation of an Autoencoder's reconstruction error throughout the network to assign relevance scores to the corresponding input feature vector.}

\section{\arno{A} Novel LRP \arno{R}ule for Autoencoders}\label{sec:lrp_and_autoencoders}

In this section, we start by introducing Layer-wise Relevance Propagation (LRP), which we leverage for \kenyu{the} explanation of Autoencoders. Then, we describe \kenyu{the} challenges of applying this technique to neural networks with \kenyu{a} reconstruction layer. Further, we briefly introduce the key concepts of \kenyu{the} Deep Taylor Decomposition method~(DTD)~\cite{dtd} that we \arno{use} to extend the LRP approach and make it applicable to Autoencoders. Finally, we present our novel LRP rule, which allows us to explain \arno{the} reconstruction error of Autoencoders.

\subsection{Layer-wise Relevance Propagation}\label{lrp:section_lrp}
Layer-wise Relevance Propagation~\cite{lrp} is an explainability method designed for neural networks to produce relevance scores for each feature of %the 
\kenyu{an} input sample~$x$. LRP assigns these scores by backward propagation from the model's output $y=f(x)$ to the input features. First, the approach assigns a relevance score to the output of the model $R_o = f(x) = y$.
\renata{Then, \arno{$R_o$ is redistributed} to the neurons from the reconstruction layer $l$, according to the LRP rule designed for the corresponding layer type. Further, relevance scores of those neurons are in turn propagated to the neurons from layer $l-1$.}
Thus, this procedure is repeated until the input layer of the network is reached. 

All LRP rules satisfy a conservation property~\cite{lrp}, which is defined by two equations. The first equation states that the sum of the relevance values received by a neuron should be equal to its \kenyu{own} relevance value:
\begin{equation}\label{lrp:eq:1}
    R_{i}^{(l)}=\sum_{k} \mathcal{R}_{i \leftarrow k}^{(l, l+1)},
\end{equation}
where $R_{i}^{(l)}$ is the relevance value assigned to neuron $i$ in layer $l$, 
$\mathcal{R}_{i \leftarrow k}^{(l, l+1)}$ is the relevance value that is distributed from neuron $k$ in layer $l+1$ to the neuron $i$ in layer $l$. The second equation states that the sum of the relevance values distributed by a neuron should be equal to its \kenyu{own} relevance value:
\begin{equation}\label{lrp:eq:2}
    R_{k}^{(l+1)}=\sum_{i} \mathcal{R}_{i \leftarrow k}^{(l, l+1)},
\end{equation}
where $ R_{k}^{(l+1)}$ is the relevance score of the neuron $k$ in layer $l+1$. 

Eq.~(\ref{lrp:eq:1}) and Eq.~(\ref{lrp:eq:2}) lead to the following layer-wise conservation property:
\begin{equation}
\label{lrp:eq:4}
 \sum_{i} R_{i}^{(l)}=\sum_{k} R_{k}^{(m)},
\end{equation}
and the following global conservation property:
 \begin{equation}
\label{lrp:eq:5}
\sum_{i} R_{i}^{(l)}=f(x),
\end{equation}
where $i$ and $k$ denote neurons' indexes in the layers $l$ and $m$ respectively. These properties define that \kenyu{the} sum of relevance scores of all neurons for a given layer is constant and equal to the relevance score, which is assigned to the output of the model $R_o=f(x)=y$.  This global conservation property \corrr{defined in \eq{lrp:eq:5}} is desirable for any explainability method that assigns relevance scores to input features~\cite{lrp}.

Using the conservation property,~\citet{lrp} define what they refer to as the basic propagation rule, which works for both convolutional and fully-connected layers, as follows:
\begin{equation}
	\label{eq:lrp_rule}
	\mathcal{R}_{i \leftarrow k}^{(l, l+1)}=R_{k}^{(l+1)} \frac{a_{i} w_{i k}}{\sum_{h \neq i} a_{h} w_{h k}},
\end{equation}
where $a_i$ is the activation value of neuron $i$, $w_{i k}$ is the weight of a link between the $i$-th and $k$-th neurons in layers $l$ and $l+1$ respectively. This rule produces an explanation that is equivalent to the gradient multiplied by the input. Further,~\citet{lrpdifflayers} propose additional rules, such as the LRP-$\epsilon$ rule to absorb the relevance values of neurons with weak activations and the LRP-$\gamma$ rule to favor the effect of positive over negative contributions. %Please 
We refer the reader to the work by~\citet{lrpdifflayers} for more information about these rules.

\subsection{Autoencoders}
\label{lrp:section_lrp_application}

Autoencoders are used in various tasks such as anomaly detection or image processing. 
%\kenyu{Comment: repeating what has been said in intro - remove?} 
These networks encode an input feature vector $x$ into a latent representation and then predict the reconstruction of the input, denoted here as $ \hat{x}$. Typically, a reconstruction loss function $e(x, \hat{x})$ is used to optimize the parameters of an Autoencoder model. One common example of such a loss function is the $L_2$ loss:
%\kenyu{Comment: should we use $e_{L1}$ and $e_{L2}$ to distinguish the two functions ?}
\begin{equation}\label{lrp:rescostfunction}
e(x, \hat{x})=\frac{1}{m} \sum_{i=1}^{m}\left(x_i - \hat{x}_i\right)^{2},
\end{equation}
where $x_i$ and $\hat{x}_i$ are the Autoencoder's input and output features, respectively, and $m$ is the dimensionality of the input feature vector $x$.  Another common example is the $L_1$ loss: 
\begin{equation}\label{lrp:rescostfunction_l1}
	e(x, \hat{x})=\frac{1}{m} \sum_{i=1}^{m}\left|x_i - \hat{x}_i\right|.
\end{equation}
We propose a method to extend the LRP explanation approach for Autoencoders to such reconstruction losses.
%explain such reconstruction losses $e(x, \hat{x})$ of Autoencoders with LRP approach.
The rule from \eq{eq:lrp_rule} is not applicable to this case, as $e(x, \hat{x})$ depends on both\corrr{,} the output and input layers of the Autoencoder. Thus, we propose a novel LRP rule that permits to propagate the reconstruction error to the Autoencoder's output layer.

\subsection{Deep Taylor Decomposition}\label{lrp:section_dtd}

Deep Taylor Decomposition (DTD)~\cite{dtd} is a similar back-propagation explainability approach that assigns relevance scores to the input features. However, while LRP rules are typically designed heuristically, DTD derives rules by using Taylor expansions.
%defines rules using Taylor expansions. 
It is interesting to note that heuristically-defined LRP rules for some types of layers have a DTD interpretation~\cite{lrpdifflayers}. Moreover, other works~\cite{lrpusecase5, lrpexample1} combine LRP and DTD rules to propagate relevance scores through an ML model. 

DTD is inspired by the divide-and-conquer paradigm, leveraging the fact that a deep neural network's function can be decomposed into a set of simpler sub-functions. These sub-functions are defined on single neurons, therefore, they can be easily expanded and decomposed using Taylor expansion. This permits the definition of a propagation rule for each neuron. Then, by aggregating multiple rules, we propagate the relevance from the output of the network to the inputs.

More formally, to obtain these decompositions we use the following two steps~\cite{dtd}:
\begin{itemize}
    \item we assume that relevance $R_j^{(l)}$ of neuron~$j$ in layer~$l$ depends solely on the set of neurons $s_{j} = \left\{{i_1, i_2, \dots}\right\}$ from the previous layer $l-1$ and, therefore, there exists a function $f_{R_j}(s_{j})= R_j^{(l)}$ ;
    \item we identify a set of neurons $\tilde{s}_{j} = \left\{{\tilde{i}_1, \tilde{i}_2, \dots}\right\}$  which are referred to as root points and serve as the starting points to compute Taylor expansion.
\end{itemize}

For any input feature vector $x$, in order to choose a root point $\tilde{s}_{j}$, we search for a set of neurons that satisfies the two following conditions~\cite{dtd}:
\begin{enumerate}
    \item $f_{R_j}(\tilde{s}_{j})=0$. This condition is necessary to obtain a decomposition that fully redistributes the relevance across the neurons 
    $\{i_1, i_2, \dots\}$;
    \item $\tilde{s}_{j}$ lies in the vicinity of $\hat{s}_{j}$ under a desired distance metric (e.g. $L_2$). Here, $\hat{s}_{j}$ is the value of neurons $s_j$ in layer $l-1$ when a sample $x$ is propagated through the network.
\end{enumerate}

Such a root point is usually obtained as a solution of an optimization problem, by minimizing the following objective:
\begin{equation}
\begin{aligned}
\label{lrp:min_cond_eq}
\tilde{s}_{j} & = \arg \min_{\xi}\|\xi -\hat{s}_j\|^{2} \\ 
%&\text { subject to } & f_{R_j}(\xi)&=0 \quad \text { and } \quad \xi \in \Xi,
& \text { s.t. } f_{R_j}(\xi) =0, \xi \in \Xi,
\end{aligned}
\end{equation}
where $\Xi$ is the input domain of $f_{R_j}$ (i.e. the set of possible neurons that influence neuron $j$).

Using the Taylor decomposition, we can then represent $R_j^{(l)}=f_{R_j}(s_{j})$ as follows:
\begin{equation}
\begin{split}
\label{lrp:eq:7}
f_{R_j}(s_{j}) & = f_{R_j}(\tilde{s}_{j})\ +
\left(\left.\frac{\partial f_{R_j}} {
	\partial {s}_{j}
}
\right|_{\tilde{s}_{j}}
\right)^{\top} \left(
s_{j} - \tilde{s}_{j}
\right)+\varepsilon_{j} \\
& = 0 + \sum_{i} \underbrace{\left.\frac{\partial f_{R_j}}{\partial i}\right|_{\tilde{s}_j} 
	\left(i-\tilde{i} \right)}_{\mathcal{R}_{i \leftarrow j}^{(l-1,l)} } +  \varepsilon_{j},
\end{split}
\end{equation}
where $\varepsilon_j$ denotes the first order Taylor residual, $|_{\tilde{s}_{j}}$ indicates that the derivative is evaluated at the root point $\tilde{s}_{j}$ and $\mathcal{R}_{i \leftarrow j}^{(l-1, l)} $ is the relevance that neuron $i$ in layer $l-1$ receives from neuron $j$ in layer $l$. \corrr{Here,} the decomposition is done only at the first-order, because the second- and higher-order terms would involve complex combinations of several neurons that propagate relevance, and, therefore, it is more challenging to derive such propagation rules~\cite{dtd}.  

By combining Eq.~(\ref{lrp:eq:4}) and Eq.~(\ref{lrp:eq:7}),
we can finally compute the relevance score of neuron $i$ in the layer $l-1$ for a chosen root point as follows:
\begin{equation}
\label{lrp:eq:9}
R_{i}^{(l-1)}=\left.\sum_{j} \frac{\partial f_{R_j}}{\partial i}\right|_{\tilde{s}_{j}} \left(i-\tilde{i}\right).
\end{equation}

\citet{dtd} suggest to choose root points based on the layer's input domain. For example, when calculating relevance scores for pixels, they propose to constrain the values of root points to be in the range between $0$ and $255$. Following Eq.~(\ref{lrp:eq:9}), other works~\cite{lrpdifflayers, lrpusecase5} describe different propagation formulas for various types of layers.

% LRP Rule for Autoencoders
\subsection{LRP Rule for Reconstruction Loss Functions}\label{sec:rule}

In this section, we describe a novel LRP rule that we can use to explain an Autoencoder's reconstruction error. This rule allows the propagation of a relevance score from the reconstruction error $R_e = e(x, \hat{x})$ to neurons from the Autoencoder's output layer. The proposed rule can be combined with other rules 
used for the remaining layers of the Autoencoder, depending on their types. Therefore, relevance scores can be seamlessly propagated all the way from the reconstruction error to the input feature vector. 

As we generate explanations for a given sample, we can assume without loss of generality that our reconstruction error~$e$ depends solely on the output neurons of the Autoencoder's output layer $\hat{x} = \{\hat{x}_1, \hat{x}_2, \dots\}$ and thus we treat the input feature vector $x$ as a constant. We then derive  
%and, thus, we treat the input feature vector $x$ as a constant. Thus, we derive 
an LRP rule for the Autoencoder's reconstruction error by decomposing the function $f_{R_e}(\hat{x}) = R_e = e(\hat{x}) $ 
%\kenyu{did you write $e(\hat{x})$ and not $e(x,\hat{x})$ on purpose ? If so, it should be consistent with Eq. (15)} using DTD.  -- added additional information to 15

In order to perform such a decomposition we need to choose a root point $\tilde{x} = \{\tilde{x}_1, \tilde{x}_2, \dots \}$ for which $f_{R_e}(\tilde{x})$ is equal to zero.
Based on the above assumptions, the only solution is the input feature vector, which is also the optimal solution for Eq.~(\ref{lrp:min_cond_eq}).

We then perform the Taylor decomposition as follows:
\begin{equation}
	\begin{aligned}
		f_{R_e}(\hat{x}) & = f_{R_e}(\tilde{x}) + \nabla f_{R_e}(\tilde{x})^{\top}(\tilde{x} - \hat{x}) + \\
		& + \frac{1}{2} (\tilde{x} - \hat{x})^{\top} H_e(\hat{x})  (\tilde{x} - \hat{x}) + \varepsilon_{e},
		\label{our:eq:taylor}
	\end{aligned}
\end{equation}
where reconstruction error $e(\tilde{x})=0$ as $\tilde{x}$ is the root point;  $H_e$ is the Hessian matrix of the reconstruction loss function $f_{R_e}$;
and $\varepsilon_{e}$ is the Taylor residual. Note that $\varepsilon_{e}=0$ for both the $L_2$ and $L_1$ loss functions introduced in~\sect{lrp:section_lrp_application}.
Below we describe in detail the derivation of the LRP rule for both $L_2$ and $L_1$ reconstruction functions.

\paragraph{$\boldsymbol{L_2}$ reconstruction function.}

\corrr{It should be noted} that we rely on the second-order Taylor decomposition for the $L_2$ loss function as its first-order derivative is equal to zero, and all the second-order terms involving multiple variables are equal to zero. Thus, we decompose $f_{R_e}(\hat{x})$ for the $L_2$ reconstruction loss from \eq{lrp:rescostfunction} as follows:
\begin{equation}
	\begin{split}
		f_{R_e}(\hat{x}) &= - \sum_{i}{\frac{2}{m}(\tilde{x}_i-%\hat{x}_i)
            \kenyu{\tilde{x}_i})(\tilde{x}_i-\hat{x}_i)}
		+\frac{1}{m}(\tilde{x}_i-\hat{x}_i)^2 \\
		%&=\sum_{i}\frac{1}{m}(\tilde{x}_i-\hat{x}_i)^2 = \sum_{i}\mathcal{R}_{i \leftarrow e}^%{(l_e, l)}, \\
            &=\sum_{i}\frac{1}{m}(\tilde{x}_i-\hat{x}_i)^2 = \sum_{i}\mathcal{R}_{i \leftarrow e}^{\kenyu{(l, l_e)}}, \\
	\end{split}
\end{equation}
where $\tilde{x}_i$ and $\hat{x}_i$ are the elements of the root point and reconstructed feature vectors correspondingly with dimension $m$; $e$ is the reconstruction error; $l$ and $l_e$ denotes the Autoencoder's output and reconstruction error layers correspondingly; 
and $\mathcal{R}_{i \leftarrow e}^{\kenyu{(l, l_e)}}$ is the relevance score that is propagated from the reconstruction error $e$ to the neuron $i$ in the Autoencoder's output layer $l$.

As each neuron $i$ in the Autoencoder's output layer receives relevance only from reconstruction error $e$, we derive the propagation rule for the $L_2$ loss as follows:
\begin{equation}
\label{lrp:eq:10}
%\mathcal{R}_{i \leftarrow e}^{(l_e, l)} = \frac{1}{m}(\tilde{x}_i-\hat{x}_i)^2
\mathcal{R}_{i \leftarrow e}^{\kenyu{(l, l_e)}} = \frac{1}{m}(\tilde{x}_i-\hat{x}_i)^2
\end{equation}

\paragraph{$\boldsymbol{L_1}$ reconstruction function.}

Similarly, we derive a propagation rule for the $L_1$ loss:
\begin{equation}\label{eq:l1_proof}
\begin{split}
f_{R_e}(\hat{x}) =\sum_{i} {\frac{1}{m}|\tilde{x}_i - \hat{x}_i|}; \\
%\mathcal{R}_{i \leftarrow e}^{(l_e, l)} =\frac{1}{m}|\tilde{x}_i - \hat{x}_i|.
\mathcal{R}_{i \leftarrow e}^{\kenyu{(l, l_e)}} =\frac{1}{m}|\tilde{x}_i - \hat{x}_i|.
\end{split}
\end{equation}
Here, the second-order term is equal to zero, while $\sum_{i}{\frac{1}{m}|\tilde{x}_i-\hat{x}_i|}$ represents the first-order term of the Taylor decomposition. More precisely, the first-order term is defined everywhere except at the singularity $\tilde{x}_i = \hat{x}_i$, where we can assume the derivative to be zero.

It is important to mention that for any input sample $x$ both of these propagation rules preserve the conservation property:
\begin{equation}
	\label{lrp:eq:11}
	\sum_{i}\mathcal{R}_{i \leftarrow e}^{(l, l_e)}  = R_{e} = e(x, \hat{x}).
\end{equation}
The proposed propagation rules for the $L_1$ and $L_2$ reconstruction loss functions allow us to extend the LRP approach to Autoencoders. In the following section we provide a detailed analysis of the proposed LRP rule by applying it to two challenging anomaly detection tasks.

\section{Experiments}\label{sec:experiments}
In this section, we describe the results of our experiments with Autoencoders for anomaly detection. We evaluate the proposed approach for Autoencoders on a SQL workload log and on image datasets. First, we assess the performance of the proposed explainability algorithm on anomalies from a SQL workload. %Since the workload is unlabelled, we introduce a self-supervised validation method based on corruption to overcome the challenge of lacking a ground truth. 
The workload is unlabelled, which makes it challenging to evaluate explainability approaches. Therefore, we introduce a self-supervised validation method based on corruption. Second, we generate and visualize explanations for images to explain the anomalous parts of objects detected by \arno{a} convolutional Autoencoder.

\subsection{Anomaly Detection on SQL Logs}
For our first experiment, we use a dataset that consists of SQL workload logs to train an anomaly detection system for database intrusion detection. The underlying model is \kenyu{a} vanilla Autoencoder-based anomaly detector that we train in an unsupervised manner with the goal of memorizing regular database activity. The dataset contains approximately $10M$ training, $5M$ validation and $10M$ test samples. The trained Autoencoder produces small reconstruction errors for test samples similar to training samples. Anomalous user behaviour results in an erroneous reconstruction of the input. An anomaly score is computed based on the $L_2$ distance between Autoencoder's input and reconstruction. The scores are normalized between $0$ and $1$.

Each dataset's sample is composed of textual and numerical features which, among others, encode various information related to the user, session, and SQL statements. Various standard embedding techniques are applied to encode the $21$ features. Explainability algorithms assign a relevance score to each of these embedded features.

We compare the three following methods:
\begin{itemize}
\item Residual explanation, which uses the $L_2$ distance between the individual original and reconstructed features as explanations, 
\item SHAP~\cite{shap}, a model-agnostic explainability approach, 
\item our approach (see Sec.~\ref{sec:lrp_and_autoencoders}) using the $L_2$ loss.
\end{itemize}

In our experiment, we use the kernel SHAP implementation with $1000$ re-evaluations of each prediction and $755$ as the background dataset size. For our approach, we use the proposed $L_2$ propagation rule for the reconstruction layer and the $z^+$ rule~\cite{lrpdifflayers} for all fully-connected layers of the Autoencoder except for the first layer, for which we apply \kenyu{the} $w^2$ rule. 

The purpose of our validation approach is two-fold. First, we quantify the performance of an explanation method to prove that the delivered explanations are satisfactory. Second, we discuss time complexity and compare the computational performance of different explainability methods.

\begin{figure*}[!h]
    \centering
    \captionsetup{justification=centering}
    \begin{tabular}{ccc}
    \includegraphics[width=0.32\textwidth]{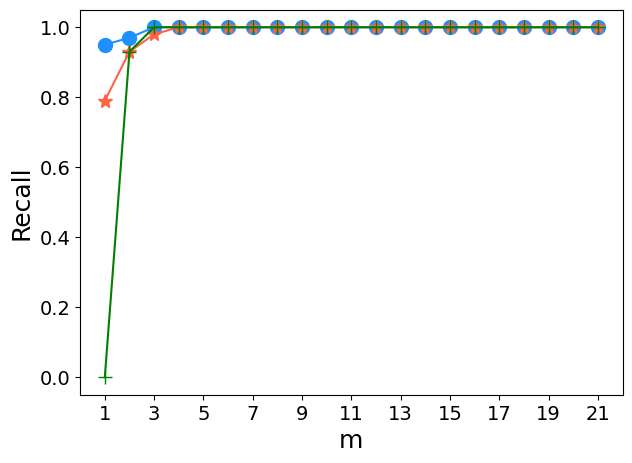} &
    \includegraphics[width=0.32\textwidth]{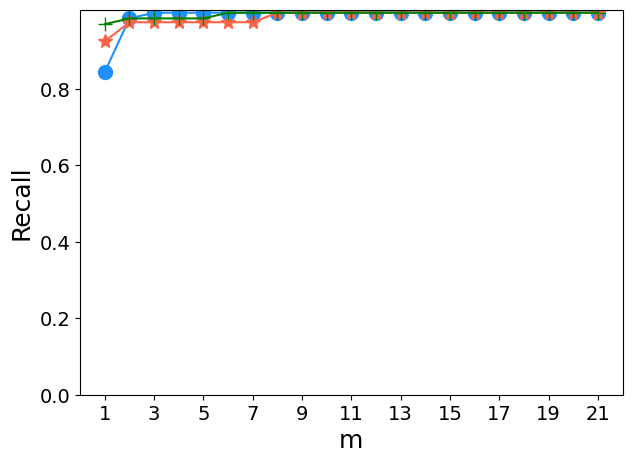} &
    \includegraphics[width=0.32\textwidth]{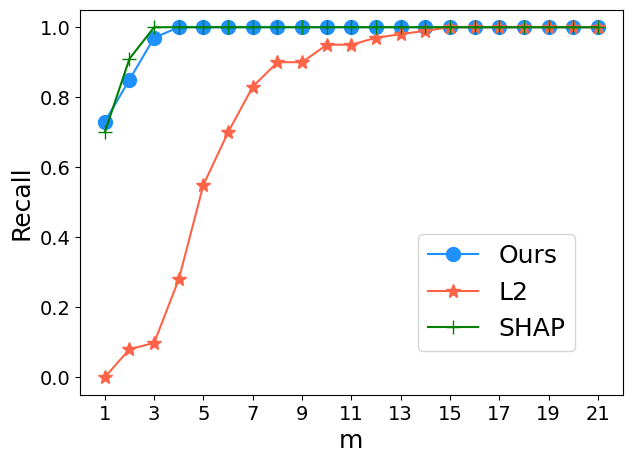} \\   
    (a)~Null corruption & (b)~Random corruption & (c)~Adversarial corruption 
    \end{tabular}
%    \caption{Resulting recall metrics for Ours (\textcolor{Cerulean}{blue $o$}), SHAP (\textcolor{Green}{green $+$}) and Residual-$L_2$ (\textcolor{RedOrange}{orange $*$}) explanation approaches applied to (a) null corruption, (b) random corruption and (c) adversarial corruption datasets\corr{, depending on $m$.}}
    \caption{Resulting recall metrics for Ours (blue $o$), SHAP (green $+$) and Residual-$L_2$ (orange $*$) explanation approaches applied to (a) null corruption, (b) random corruption and (c) adversarial corruption datasets.}
    \label{fig:precision_recall_sql}
\end{figure*}

\paragraph{Quantitative comparison.}
We quantify the performance of an attribution-based explanation method through a validation method based on corruption. Our approach consists in modifying one input feature of a clean and randomly chosen sample. The modification changes the feature's value in such a way that the Autoencoder produces a high anomaly score for the modified sample. Thus, we obtain a ground truth that indicates the feature causing the high anomaly score of the given sample, and permits the computation of a validation metric to compare different explainability approaches.

For the input feature modification we use the following three corruption strategies: null, random and adversarial. The null corruption method changes the feature's value to 0; the random corruption approach modifies the feature's value to a random value sampled from a uniform distribution between 0 and 1 (that is, the same range as the initial feature values). Finally, the adversarial corruption method updates the feature's value in such a way that reconstruction of the given feature does not increase while the the reconstruction error of other features increases. Further details are given in the Appendix. 

Then, we generate $K=100$ anomalous samples with the aforementioned corruptions. All these generated samples have anomaly scores greater than a threshold $T$. We set $T$ = 0.3, $T$ = 0.5 and $T$ = 0.3 for the adversarial, random and null corruptions, respectively. We have chosen these values empirically, based on the difficulty for the corruption method to produce datapoints with an anomaly score exceeding the given threshold. In practice, we expect explanation methods to achieve better performance when $T$ is large, as the corrupted feature's contribution to the anomaly becomes large. Conversely, when $T$ is lower, the validation approach leads to assessing an explanation method's sensitivity to identifying a corrupted feature with a lower contribution to the anomaly.

To compare different approaches we use a recall metric that we calculate as follows. We define an explanation to be correct when a corrupted feature is among the $m \in \{1, \dots, M\}$ features with the highest anomaly scores, where $M$ is the number of features. Otherwise, we define an explanation to be incorrect. We calculate the recall metric based on generated validation samples $recall = N_{+} / (N_{+} + N_{-})$, where $N_{+}$ and $N_{-}$ denote the number of correct and incorrect explanations respectively.

Figure~\ref{fig:precision_recall_sql} illustrates the recall metric defined above for \kenyu{the} null, random and adversarial corruption validation datasets correspondingly. The experiment shows that the Residual explanation achieves \arno{a} good validation score on \kenyu{the} null and random corruption datasets. However, \arno{the performance} on the more challenging adversarial corruption dataset \arno{is poor}. On the other hand, SHAP and our method achieve good performance on all datasets. This shows that our proposed approach and SHAP are more generic and succeed in explaining a broader range of anomalies.

\paragraph{Time Complexity.}
Even though SHAP produces accurate results, our method delivers explanations of comparable quality with substantially lower time complexity. This is explained by the fact that our approach only requires one forward and backward pass to compute relevance scores, while SHAP requires a forward pass for each generated perturbation and for each re-evaluation. In our experiments, using the parameters described at the beginning of this Section, the execution time required to compute explanations is three to four orders of magnitude faster with our method compared to SHAP.

%Table~\ref{table:time} illustrates the preceding claim by quantifying \kenyu{the} execution time to generate explanations using each of the two methods.

%\begin{table}[!t]
%\centering
%\caption{Execution time required to compute explanations for 30 datapoints}
%\begin{tabularx}{\linewidth}{Xr} 
%\toprule
%Method & Execution Time (s) \\ 
%\midrule
%SHAP   & $\sim 2000$\\ 
%Ours  & \textbf{$\sim$ 0.5} \\ 
%\bottomrule
%\end{tabularx}
%\label{table:time}
%\end{table}

We conclude that SHAP and the proposed approach produce more accurate results compared to \kenyu{the} Residual explanation. Nonetheless, as our method is several orders of magnitude faster than SHAP, our approach is more suitable for time-sensitive applications.

\subsection{Anomaly Detection on Image Dataset}\label{sec:image_anomaly_detection}

For our visual anomaly detection experiments we use images from the MVTec dataset~\cite{anomaly_image_dataset}, which contains several objects with various types of damage. \renata{For each of these objects we train a convolutional Autoencoder model as suggested by~\citet{autoencoder_on_images}. The training set consists of images of non-damaged objects and \kenyu{the} test set contains images of damaged objects. In this case \arno{the} Autoencoder is expected to show high reconstruction error for the damaged parts of objects in the test set. In this experiment, we provide explanations for this reconstruction error and compare them with ground-truth images of damaged areas that are also provided in the dataset.}

\renata{We} convert each image to gray-scale, apply a Gaussian filter with kernel size $3$ and $\sigma=0.5$ and resize to $128 \times 128$ pixels. To avoid overfitting, we use $10\%$ of the training dataset as validation. We also preprocess the training and validation data by applying random rotations and flips to the images. Using this augmentation method, we generate $10000$ training samples and $1000$ validation samples for each class of objects. We use a similar architecture as the one proposed by~\citet{autoencoder_on_images}, with the following modifications: 
\begin{itemize}
\item all leaky ReLU activations are replaced with ReLU activations;
\item convolutional kernels sizes are modified.
\end{itemize}
\corrr{Appendix~\ref{appendix:adverserial_corruption} provides more details about the exact model architecture}.

To evaluate our explanations, we rely on the precision-recall metric. We compute this metric separately for the various damaged object groups. For each group we set anomaly thresholds: $T=(t_1, \dots,  t_i, \dots, t_{1000})$ in range $(\min(R_{j}), \max(R_{j}))$, where $\min(R_{j})$ and $\max(R_{j})$ are minimal and maximal pixel relevance values across all pixels and all images of the given group $j$. Any pixel that is assigned a relevance value higher than $t_i$ is considered \arno{as} damaged. For each threshold $t_i$, we calculate precision and recall values by taking into account all pixels that have relevance values higher than $t_i$. Further, we compute the average precision (AP) metric by calculating the area under the precision-recall curve.

 \begin{table}[!ht]
 \centering
 \small
 \caption{\renata{AP metric for several object classes from the MVTec dataset~\cite{anomaly_image_dataset}. Here we compare the $L_1$ and $L_2$ Residual explanation methods with the proposed approach for $L_1$ and $L_2$ reconstruction functions.}}
     \begin{tabularx}{\linewidth}{cXcc|cc} 
     % \cline{1-4}
     \toprule
     \multirow{2}{*}{Object} & \multirow{2}{*}{Damage Type} & \multicolumn{2}{c|}{Residual} & \multicolumn{2}{c}{Ours} \\ 
      &  & $L_1$ & $L_2$ & $L_1$ & $L_2$ \\ 
     \midrule
     \multirow{4}{*}{\rotatebox{90}{Transistor}} & Bent Lead & \textbf{0.104} & \textbf{0.103} & 0.027 & {0.028} \\
     & Cut Lead & \textbf{0.082} & 0.079 & 0.038 & {0.040} \\
     & Damaged Case & {0.061}  & 0.058 & \textbf{0.122} & 0.103 \\
     & Misplaced & {0.494} & 0.498 & \textbf{0.888} & {0.867} \\
     \midrule
     \multirow{3}{*}{\rotatebox{90}{Bottle}} & Broken Large & {0.305} & 0.298 & \textbf{0.366} & \textbf{0.366} \\
     & Broken Small & {0.206} & 0.202 & \textbf{0.340}  & 0.321 \\
     & Contamination & {0.148} & 0.155 & \textbf{0.369} &{0.263} \\
     \midrule
     \multirow{3}{*}{\rotatebox{90}{Hazelnut}} & Crack & \textbf{0.478} & 0.475 & 0.317 & {0.295} \\
     & Hole & 0.455 & \textbf{0.457} & 0.240 &{0.218} \\
     \vspace{1mm}
     & Print & \textbf{0.661} & \textbf{0.660} & 0.557 &{0.514} \\
     \midrule
     %Overall & & {0.30} & 0.30 & \textbf{0.33} & 0.30 \\
     Overall & & 0.298 & 0.298 & \textbf{0.328} & 0.304 \\
     \bottomrule
     \end{tabularx}
 \label{table:MAP_metrics}
 \end{table}

%5
\begin{figure*}[!h]
    \centering
    \begin{tabular}{cccccccc}
        \includegraphics[width=0.11\textwidth]{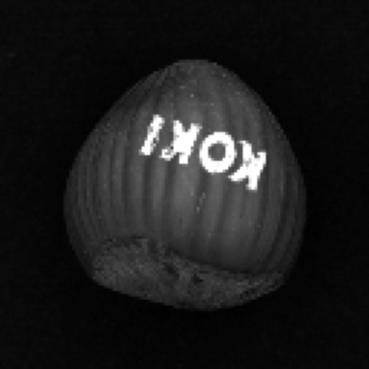}&
        \hspace{-0.25cm}\includegraphics[width=0.11\textwidth]{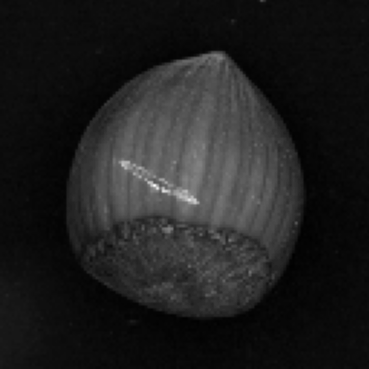}&
        \hspace{-0.25cm}\includegraphics[width=0.11\textwidth]{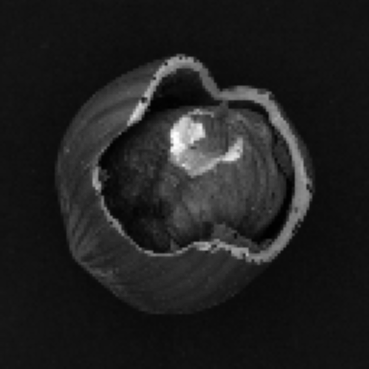}&
        \hspace{-0.25cm}\includegraphics[width=0.11\textwidth]{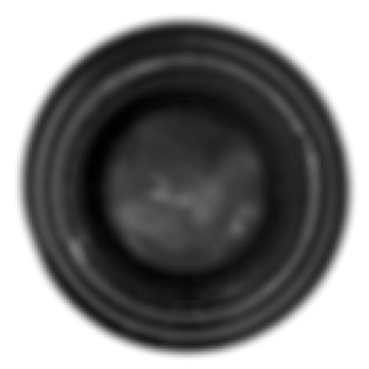}&
        \hspace{-0.25cm}\includegraphics[width=0.11\textwidth]{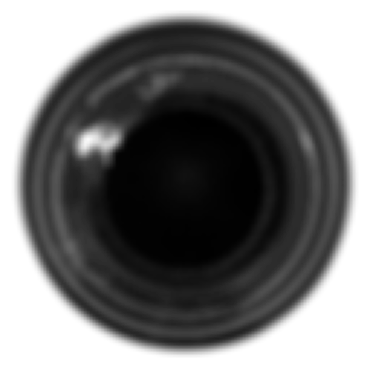}&
        \hspace{-0.25cm}\includegraphics[width=0.11\textwidth]{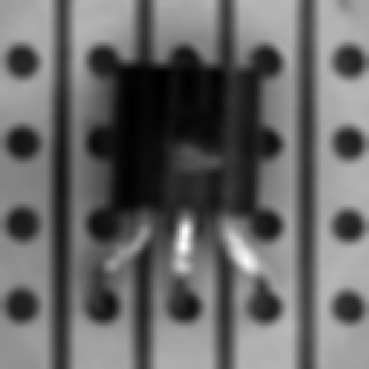}&
        % \hspace{-0.25cm}\includegraphics[width=0.11\textwidth]{figures/ex10_original.png}&
        % \hspace{-0.25cm}\includegraphics[width=0.11\textwidth]{figures/ex4_original.png}
        \hspace{-0.25cm}\includegraphics[width=0.11\textwidth]{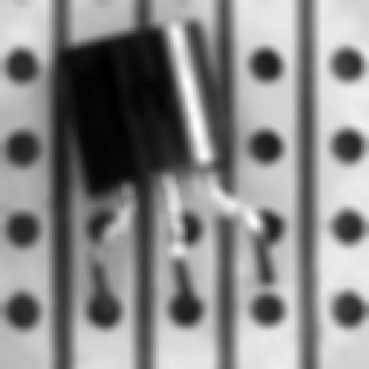}&
        \hspace{-0.25cm}\includegraphics[width=0.11\textwidth]{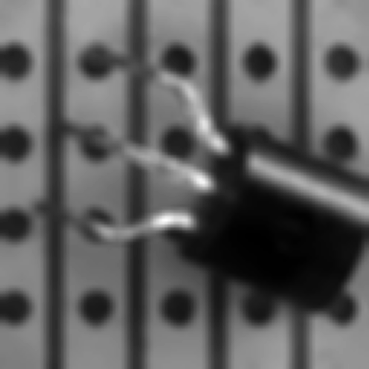}
        \\
        \multicolumn{8}{c}{(a) Original} 
        \\
        \includegraphics[width=0.11\textwidth]{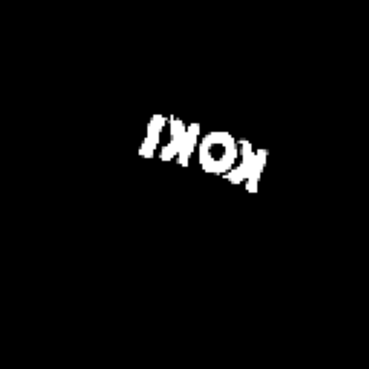}&
        \hspace{-0.25cm}\includegraphics[width=0.11\textwidth]{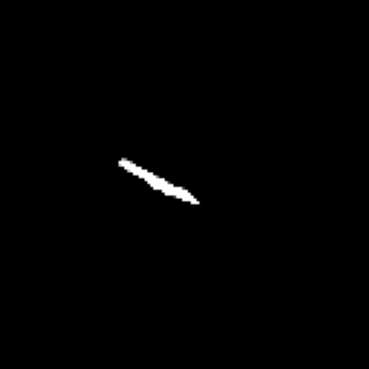}&
        \hspace{-0.25cm}\includegraphics[width=0.11\textwidth]{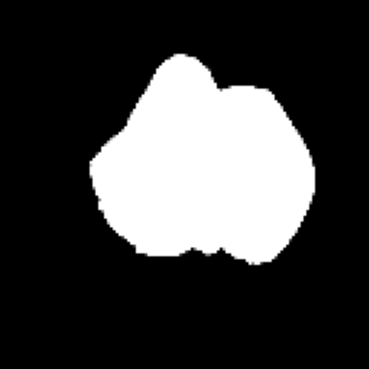}&
        \hspace{-0.25cm}\includegraphics[width=0.11\textwidth]{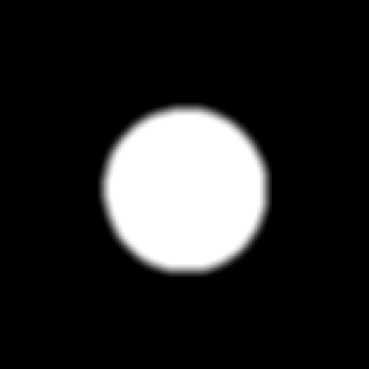}&
        \hspace{-0.25cm}\includegraphics[width=0.11\textwidth]{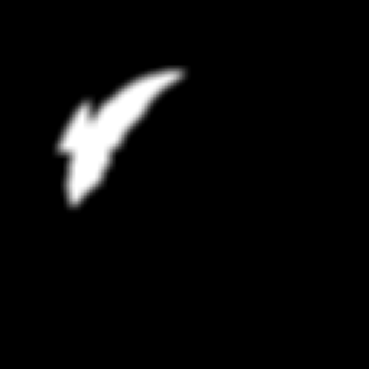}&
        \hspace{-0.25cm}\includegraphics[width=0.11\textwidth]{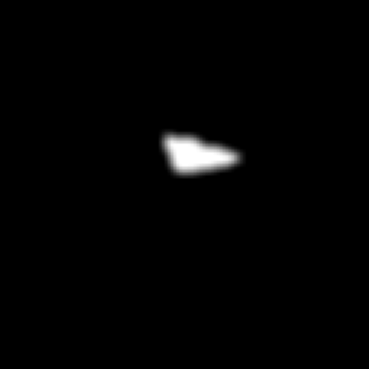}&
        % \hspace{-0.25cm}\includegraphics[width=0.11\textwidth]{figures/ex4_g_.png}&
        % \hspace{-0.25cm}\includegraphics[width=0.11\textwidth]{figures/ex10_g_.png}
        \hspace{-0.25cm}\includegraphics[width=0.11\textwidth]{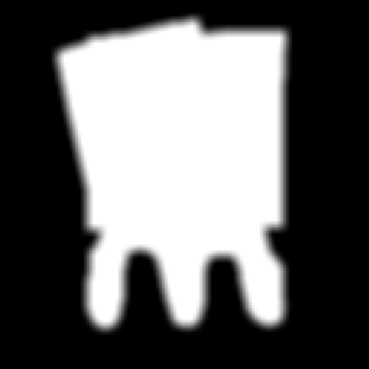}&
        \hspace{-0.25cm}\includegraphics[width=0.11\textwidth]{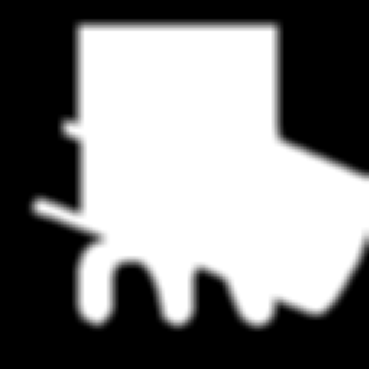}
        \\
        \multicolumn{8}{c}{(b) Ground-truth} 
        \\        
        \includegraphics[width=0.11\textwidth]{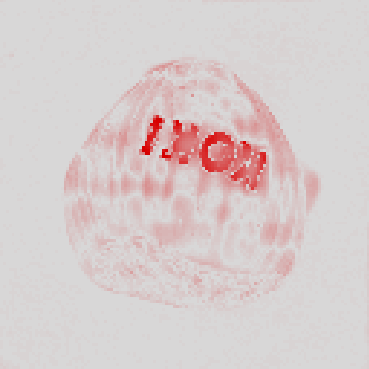}&
        \hspace{-0.25cm}\includegraphics[width=0.11\textwidth]{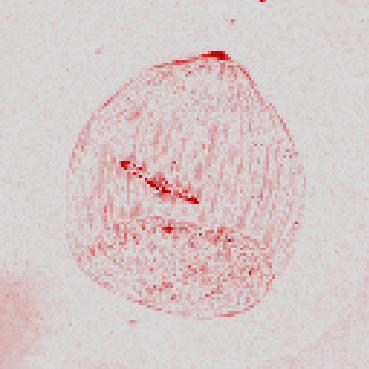}&
        \hspace{-0.25cm}\includegraphics[width=0.11\textwidth]{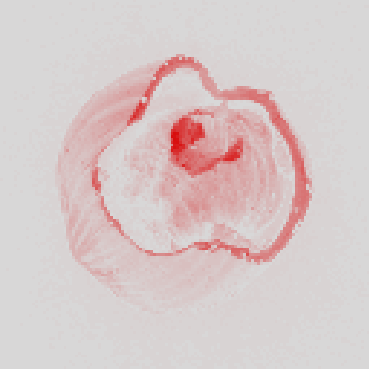}&
        \hspace{-0.25cm}\includegraphics[width=0.11\textwidth]{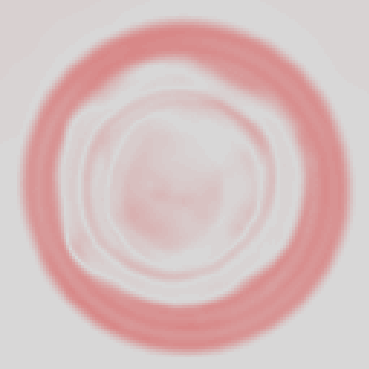}&        
        \hspace{-0.25cm}\includegraphics[width=0.11\textwidth]{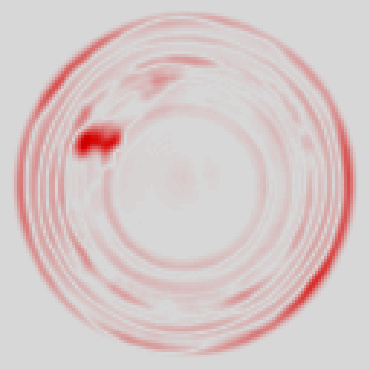}&
        \hspace{-0.25cm}\includegraphics[width=0.11\textwidth]{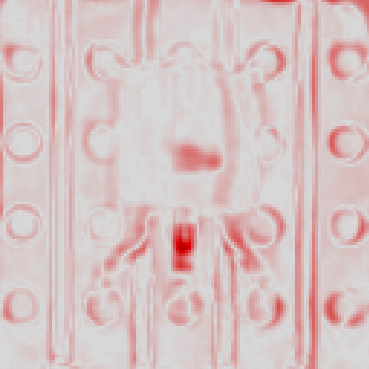}&
        % \hspace{-0.25cm}\includegraphics[width=0.11\textwidth]{figures/ex10_l2.png}&
        % \hspace{-0.25cm}\includegraphics[width=0.11\textwidth]{figures/ex4_l2.png}
        \hspace{-0.25cm}\includegraphics[width=0.11\textwidth]{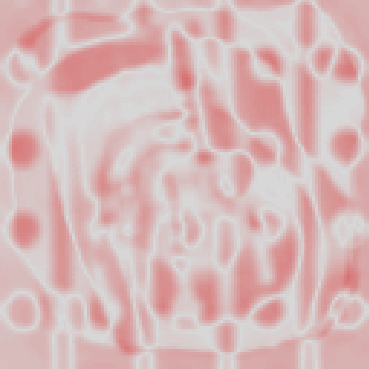}&
        \hspace{-0.25cm}\includegraphics[width=0.11\textwidth]{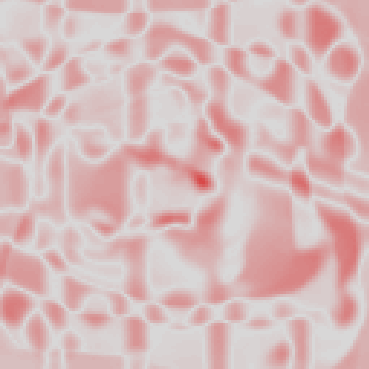}
        \\
        \multicolumn{8}{c}{(c) Residual-$L_1$} 
        \\
        \includegraphics[width=0.11\textwidth]{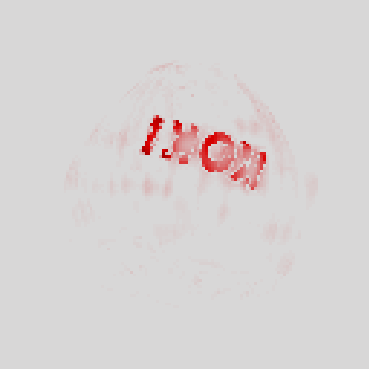}&
        \hspace{-0.25cm}\includegraphics[width=0.11\textwidth]{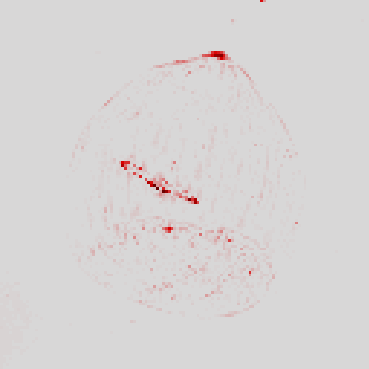}&
        \hspace{-0.25cm}\includegraphics[width=0.11\textwidth]{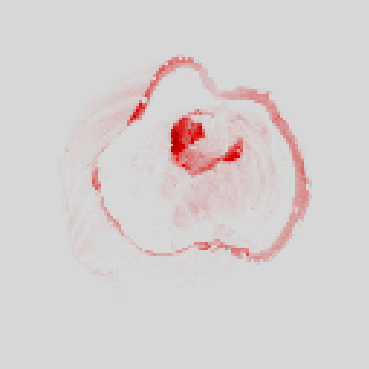}&
        \hspace{-0.25cm}\includegraphics[width=0.11\textwidth]{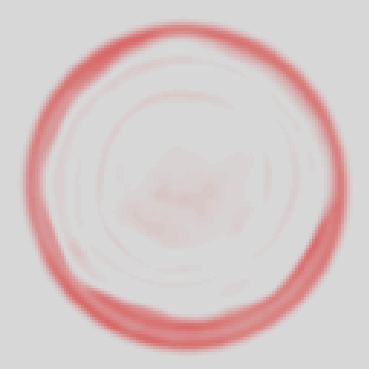}&
        \hspace{-0.25cm}\includegraphics[width=0.11\textwidth]{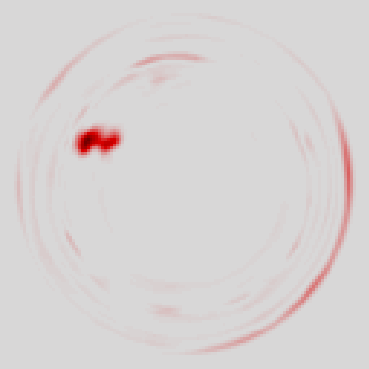}&
        \hspace{-0.25cm}\includegraphics[width=0.11\textwidth]{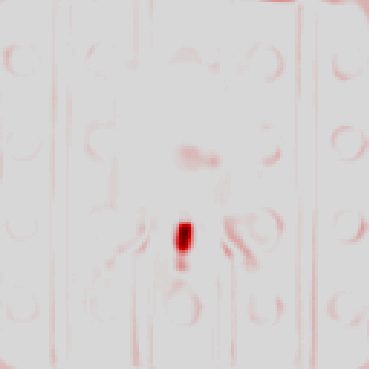}&
        % \hspace{-0.25cm}\includegraphics[width=0.11\textwidth]{figures/ex4_l_.png}&
        % \hspace{-0.25cm}\includegraphics[width=0.11\textwidth]{figures/ex10_l_.png}
        \hspace{-0.25cm}\includegraphics[width=0.11\textwidth]{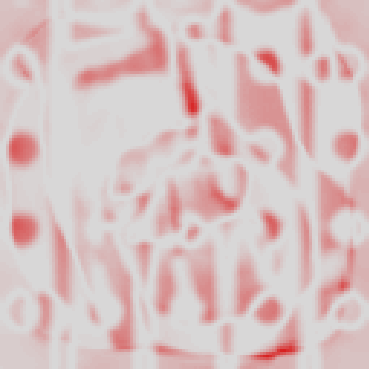}&
        \hspace{-0.25cm}\includegraphics[width=0.11\textwidth]{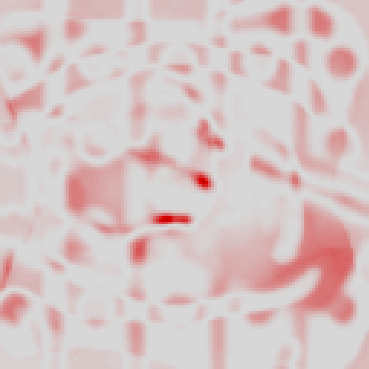}
        \\
        \multicolumn{8}{c}{(d) Residual-$L_2$} 
        \\
        \includegraphics[width=0.11\textwidth]{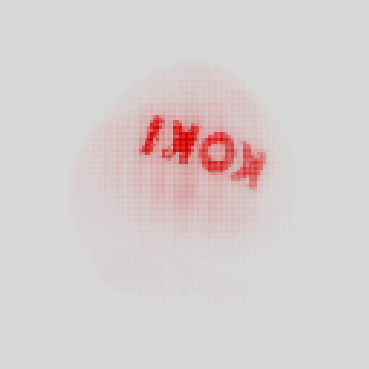}&
        \hspace{-0.25cm}\includegraphics[width=0.11\textwidth]{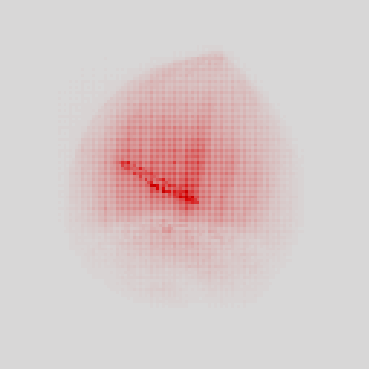}&
        \hspace{-0.25cm}\includegraphics[width=0.11\textwidth]{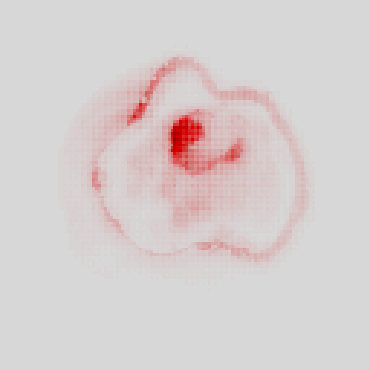}&
        \hspace{-0.25cm}\includegraphics[width=0.11\textwidth]{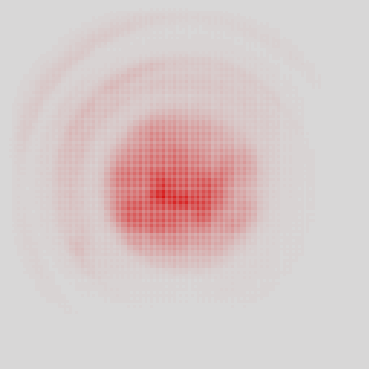}&
        \hspace{-0.25cm}\includegraphics[width=0.11\textwidth]{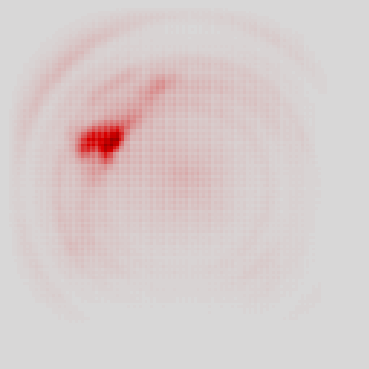}&
        \hspace{-0.25cm}\includegraphics[width=0.11\textwidth]{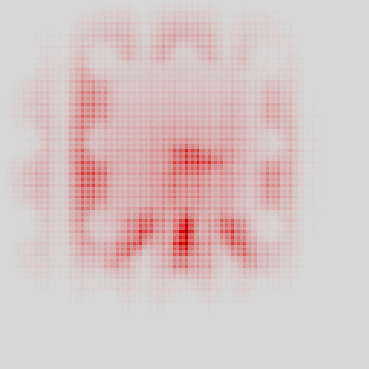}&
        % \hspace{-0.25cm}\includegraphics[width=0.11\textwidth]{figures/ex4_z1_.png}&
        % \hspace{-0.25cm}\includegraphics[width=0.11\textwidth]{figures/ex10_z1_.png}
        \hspace{-0.25cm}\includegraphics[width=0.11\textwidth]{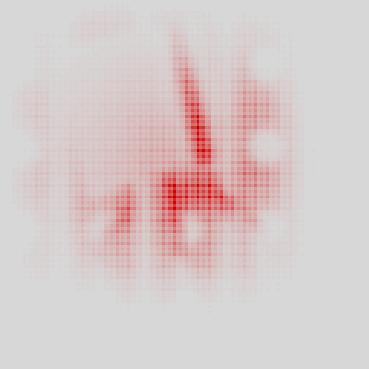}&
        \hspace{-0.25cm}\includegraphics[width=0.11\textwidth]{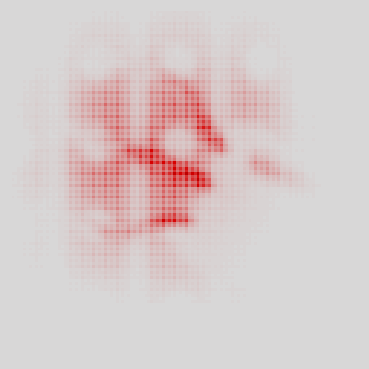}
        \\
        \multicolumn{8}{c}{(e) Ours-$L_1$}         
        \\
        \includegraphics[width=0.11\textwidth]{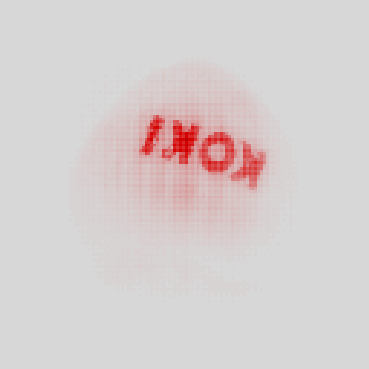}&
        \hspace{-0.25cm}\includegraphics[width=0.11\textwidth]{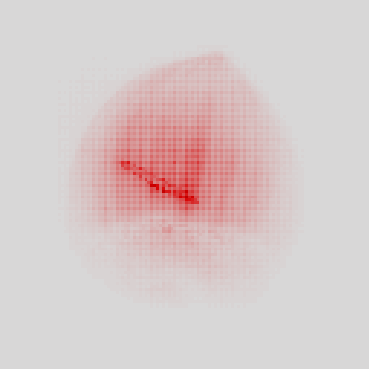}&
        \hspace{-0.25cm}\includegraphics[width=0.11\textwidth]{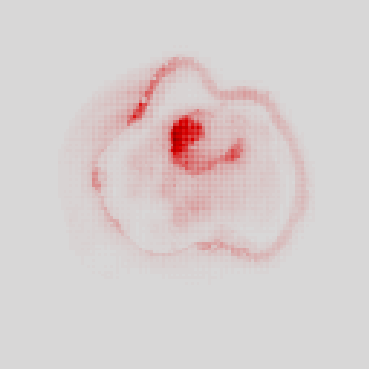}&
        \hspace{-0.25cm}\includegraphics[width=0.11\textwidth]{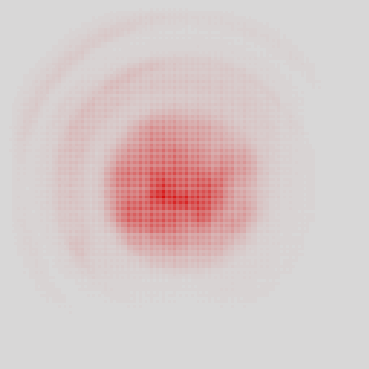}&
        \hspace{-0.25cm}\includegraphics[width=0.11\textwidth]{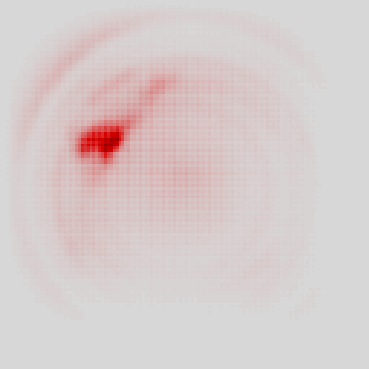}&
        \hspace{-0.25cm}\includegraphics[width=0.11\textwidth]{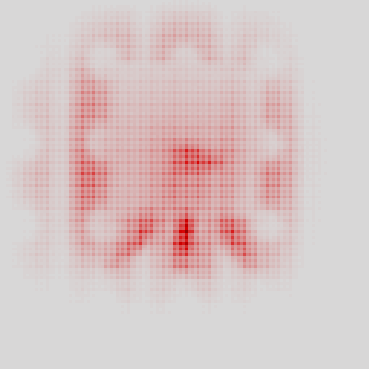}&
        % \hspace{-0.25cm}\includegraphics[width=0.11\textwidth]{figures/ex10_z.png}&
        % \hspace{-0.25cm}\includegraphics[width=0.11\textwidth]{figures/ex4_z.png}
        \hspace{-0.25cm}\includegraphics[width=0.11\textwidth]{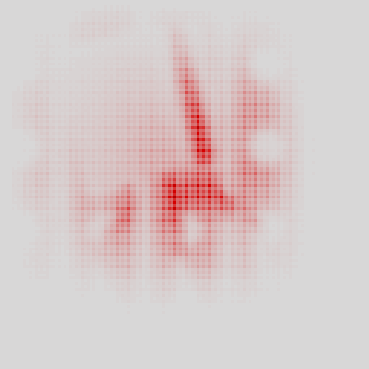}&
        \hspace{-0.25cm}\includegraphics[width=0.11\textwidth]{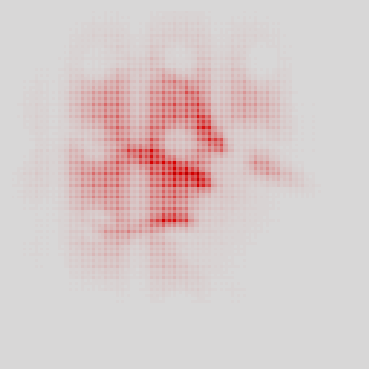}
        \\
        \multicolumn{8}{c}{(f) Ours-$L_2$} 
        \\
    \end{tabular}
    \caption{Explanations produced for images of damaged objects from the MVTec dataset~\cite{anomaly_image_dataset}. The figure illustrates the (a) original image; (b) ground truth damaged area; Autoencoder's explanations produced by the (c) baseline Residual-$L_1$ and (d) Residual-$L_2$ explainability methods and (e,f) results of the proposed LRP-based approach. For our approach, we show explanations generated by propagation of (e) $L_1$ and (f) $L_2$ reconstruction losses. As we can see, the baseline methods focus on the borders of the object, while our approach focuses the attention on the damaged areas. Additional examples are provided in the Appendix~\ref{sec:examples}.}
    \label{fig:visual_result}
\end{figure*}

We rely on two baseline methods to evaluate \kenyu{the} performance of our convolutional Autoencoder's explanation approach, which we denote as Residual-$L_1$ and Residual-$L_2$. These methods calculate reconstruction error as $(x - \hat{x})^2$ and $|x - \hat{x}|$ respectively. Similarly, for our approach we investigate the following two settings, when the reconstruction error is computed using either $L_1$ or $L_2$ losses and propagated using \kenyu{the} LRP-rules \kenyu{from} \eq{lrp:eq:10} and \eq{eq:l1_proof} respectively. We refer to these methods as Ours-$L_1$ and Ours-$L_2$.
In both these methods we rely on the $z^+$-rule for the relevance propagation through convolutional layers, and $z$-box-rule for the first layer.

Table~\ref{table:MAP_metrics} shows a comparison of AP for several object classes from the MVTec dataset~\cite{anomaly_image_dataset}. On average, our approaches produce higher scores compared to the baseline methods. We also notice that some damages are harder to explain, e.g. the transistor object class with the cut lead damage. The main reason is that the Autoencoder model is not able to reach good reconstruction accuracy, which consequently lowers the performance for all explainability approaches.

Furthermore, Figure~\ref{fig:visual_result} illustrates explanations that are produced by the proposed approach and baseline method. We notice that our explanations focus on the important part of the object and its neighbouring area that belongs to the object. Also, it does not assign much importance to the background pixels. In contrast, the Residual explanations highlight various parts of images, which are primarily borders and are not necessarily relevant to the damaged part of the object. Additional examples to highlight the preceding statement are provided in the Appendix~\ref{fig:appendix_result}.

Finally, we analyze images, on which Residual explanations outperform our approach with respect to the AP metric. We notice that \corrr{the} score does not always reflect \kenyu{the} quality of produced explanations. For example, the samples of objects with the hazelnut class, which is depicted by Figure~\ref{fig:visual_result}, have pixels with high relevance scores outside of the ground truth area. However, all of those pixels are located next to the correct damaged parts of the object. In contrast, the Residual explanation highlights some object borders, which are not relevant to the damages, but the amount of highlighted pixels is smaller outside the ground truth area, which results in a high AP metric. Therefore, \corrr{the} AP metric might not be ideal for evaluation of explainability approaches. While developing an appropriate metric is out of the scope of this work, it is an interesting direction for future research.

\section{Conclusion}\label{sec:conclusion}
In this work, we have proposed a principled way to extend an LRP explainability technique for Autoencoders using Deep Taylor Decomposition~\cite{dtd}. Furthermore, we have suggested a self-supervised validation technique for attribution-based explanation methods by leveraging various corruption methods. Finally, our experiments show that the proposed method outperforms the Residual explanation baseline method and shows comparable performance to model-agnostic approaches such as SHAP~\cite{shap}, while being several orders of magnitude faster.

\bibliography{references}
\bibliographystyle{icml2023}

%%%%%%%%%%%%%%%%%%%%%%%%%%%%%%%%%%%%%%%%%%%%%%%%%%%%%%%%%%%%%%%%%%%%%%%%%%%%%%%
%%%%%%%%%%%%%%%%%%%%%%%%%%%%%%%%%%%%%%%%%%%%%%%%%%%%%%%%%%%%%%%%%%%%%%%%%%%%%%%
% APPENDIX
%%%%%%%%%%%%%%%%%%%%%%%%%%%%%%%%%%%%%%%%%%%%%%%%%%%%%%%%%%%%%%%%%%%%%%%%%%%%%%%
%%%%%%%%%%%%%%%%%%%%%%%%%%%%%%%%%%%%%%%%%%%%%%%%%%%%%%%%%%%%%%%%%%%%%%%%%%%%%%%
\newpage
\appendix
\onecolumn
\section{Adverserial Corruption}\label{appendix:adverserial_corruption}

A commonly used approach to generating adversarial examples is the Iterative Fast Gradient Sign Method (I-FGSM) \cite{iterative_adv}, which uses the Fast Gradient Sign Method (FGSM) \cite{adversarial_example}. I-FGSM applies FGSM multiple times with a small step size to maximize the output of a loss function $L$ given $x^{adv}$, and clips values of intermediate results after each step to ensure that they are in an $\epsilon$-neighbourhood of the original datapoint $x$:

\begin{gather}
\label{validation_2:ifgsm1}
{x}_{0}^{a d v}={x}\\
\label{validation_2:ifgsm2}
\quad {x}_{N+1}^{a d v}=\operatorname{Clip}_{x, \epsilon}\left({x}_{N}^{a d v}+\alpha \operatorname{sign}\left(\nabla_{x} L\left({x}_{N}^{a d v}\right)\right)\right)
\end{gather}

where $\operatorname{Clip}_{x, \epsilon}$ indicates the resulting datapoint is clipped within the $\epsilon$-ball of the original datapoint $x$, $N$ is the iteration number, $\alpha$ is the step size, and $L\left({x}_{N}^{a d v}\right)$ is the model's loss on ${x}_{N}^{a d v}$.

Adversarial corruption extends I-FGSM to generate anomalous datapoints. Notably, the method corrupts one randomly selected feature of a non-anomalous datapoint. By definition, such a datapoint has a low reconstruction error. The corruption aims to maximize the reconstruction error of the resulting datapoint for all features except the corrupted one. For the latter, the error is minimized. To that end, we make the following adaptations to the I-FSGM:
%The corruption aims to maximize the reconstruction error of the resulting datapoint while the corrupted feature's contribution to the reconstruction error is minimized.

\begin{enumerate}
\item We define the loss function to be maximized in this context. We define $x_i$ as the $i$-th feature of $x$, and $x_{c}$ the feature to be corrupted. Furthermore, we define feature $x_i$'s reconstruction error as $\phi(x,i) = \left(x_{i} - \hat{x_{i}}\right)^{2}$, where $\hat{x_{i}}$  is the reconstruction of feature $x_i$. The goal is to minimize $x_{c}$'s reconstruction error, while maximizing the ones from $x_i$, for all $i \neq c$. This translates as maximizing the following loss function:
\begin{equation}
\label{validation_2:custom_loss_function}
r_{adv}(x,c)=\frac{1}{m} \left[ \left( \sum_{i=1 ; i \neq c}^{m} \phi(x,i) \right) - \theta* \phi(x,c) \right],
\end{equation}
where $x$ designates the Autoencoder's input feature vector, $c$ is the index of the feature to be corrupted, $m$ is the size of $x$, $\theta$ is the weight controlling the extent to which the $x_{c}$'s reconstruction error should be minimized relatively to the global optimization objective.
\item We emphasize the fact that the adversarial corruption should only modify $x_{c}$, as the resulting datapoint's anomaly should be caused by $x_c$. Therefore, the updates on $x$ should only concern $x_{c}$.
\item We do not require the adversarial example to be in the neighbourhood of the original datapoint $x$.
\end{enumerate}

Applying the above points, we reduce Eq.~\ref{validation_2:ifgsm1} and Eq.~\ref{validation_2:ifgsm2} to formulate the following adversarial optimization:
\begin{gather}
\label{validation_2:ifgsm_no_clip1}
{x_{{0}}^{a d v}}={x},\\ 
%{x_{c_{0}}^{a d v}}={x_{c}},\\ 
\label{validation_2:ifgsm_no_clip2}
\quad {x_{c_{N+1}}^{a d v}}={x_{c_{N}}^{a d v}}+\alpha \operatorname{sign}\left(\nabla_{x_{c}}r_{adv}\left({x}_{N}^{a d v}, c\right)\right)
%{x}_{0}^{a d v}={x}, \quad {x}_{N+1}^{a d v}={x}_{N}^{a d v}+\alpha \operatorname{sign}\left(\nabla_{x}\ r_{adv}\left({x}_{N}^{a d v}\right)\right)\end{gather},
\end{gather}
where ${x}_{N}^{a d v}$ corresponds to the datapoint $x^{adv}$ after $N$ update steps. Since the updates only concern $x_c$, ${x}_{N}^{a d v}$ is identical to ${x}_{0}^{a d v}$ for all features except $x_c$, i.e. ${x}_{i_{N}}^{a d v}$ = ${x}_{i{0}}^{a d v}$, for all $i \neq c$.

We refer to the above adversarial optimization, which produces $x^{adv}$ given $x$, as the adversarial corruption.

Furthermore, to improve results with adversarial corruption we introduce a step size updater, which halves the value of $\alpha$ every $k$ updates, if the two following conditions are both met:

\begin{enumerate}
\item The reconstruction error of the corrupted feature $\phi(x,c)$ did not decrease;
\item The sum of the reconstruction errors of the non-corrupted features $ \sum_{i=1 ; i \neq c}^{m} \phi(x,i) $ did not increase. 
\end{enumerate}

Lastly, we strongly recommend performing the adversarial optimization on a datapoint which has been subject to random corruption. This enables the initial anomaly score to reach a desired threshold, and experiments show that optimization starting from that regime leads to better results. In that case, the adversarial optimization should be done on the same corrupted feature as the one subjected to random corruption.

\section{Image Anomaly Detection}
\subsection{Autoencoder Architecture}\label{sec:architecture}
We use a modified version of a convolutional Autoencoder architecture defined in \citet{autoencoder_on_images} to run the experiments described in Sec. \ref{sec:image_anomaly_detection}. 

\begin{table}[H]
\centering
    \begin{tabularx}{\linewidth}{lllllll} 
    % \cline{1-4}
    \toprule
    Layer & Output Size & Kernel & Stride & Padding & Activation & Scaling Factor\\
    \midrule
    Input & 128 x 128 x 1 & - & - & - & - & -\\
    Conv 1 & 62 x 62 x 32 & 5 & 2 & 0 & ReLu & - \\
    Conv 2 & 30 x 30 x 32 & 3 & 2 & 0 & ReLu & -\\
    Conv 3 & 28 x 28 x 32 & 3 & 1 & 0 & ReLu & -\\
    Conv 4 & 12 x 12 x 64 & 5 & 2 & 0 & ReLu & -\\
    Conv 5 & 10 x 10 x 64 & 3 & 1 & 0 & ReLu & -\\
    Conv 6 & 4 x 4 x 128 & 3 & 2 & 0 & ReLu & -\\
    Conv 7 & 2 x 2 x 512 & 3 & 1 & 0 & ReLu & -\\
    \midrule
    Upsample & 6 x 6 x 512 & - & - & - & - & 3.0\\
    Conv 8 & 4 x 4 x 128 & 3 & 1 & 0 & ReLu & -\\
    Upsample & 12 x 12 x 128 & - & - & - & - & 3.0\\
    Conv 9 & 10 x 10 x 64 & 3 & 1 & 0 & ReLu & -\\
    Upsample & 20 x 20 x 64 & - & - & - & - & 3.0\\
    Conv 10 & 18 x 18 x 64 & 3 & 1 & 0 & ReLu & -\\
    Upsample & 36 x 36 x 64 & - & - & - & - & 2.0\\
    Conv 11 & 34 x 34 x 32 & 3 & 1 & 0 & ReLu & -\\
    Upsample & 68 x 68 x 32 & - & - & - & - & 2.0\\
    Conv 12 & 66 x 66 x 32 & 3 & 1 & 0 & ReLu & -\\
    Upsample & 132 x 132 x 32 & - & - & - & - & 2.0\\
    Conv 13 & 130 x 130 x 32 & 3 & 1 & 0 & ReLu & -\\
    Conv 14 & 128 x 128 x 1 & 3 & 1 & 0 & ReLu & -\\
    \bottomrule
    \end{tabularx}
    \caption{Convolutional Autoencoder architecture with 14 layers for unsupervised object anomaly detection. Upsampling uses nearest pixel interpolation.}
\label{table:autoencoder}
\end{table}

\newpage
\subsection{Examples}\label{sec:examples}
\begin{figure*}[!h]
    \centering
    \begin{tabular}{cccccccc}
        \includegraphics[width=0.11\textwidth]{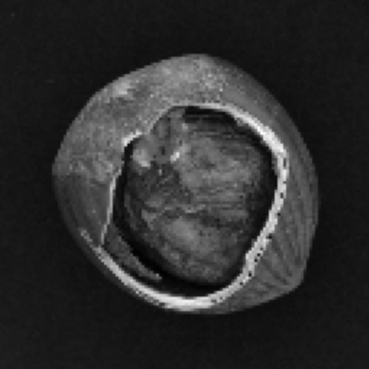}&
        \hspace{-0.25cm}\includegraphics[width=0.11\textwidth]{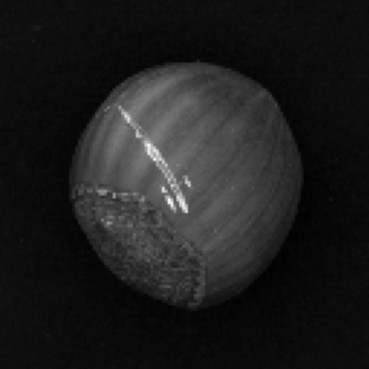}&
        \hspace{-0.25cm}\includegraphics[width=0.11\textwidth]{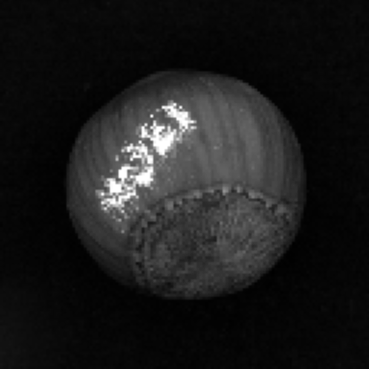}&
        \hspace{-0.25cm}\includegraphics[width=0.11\textwidth]{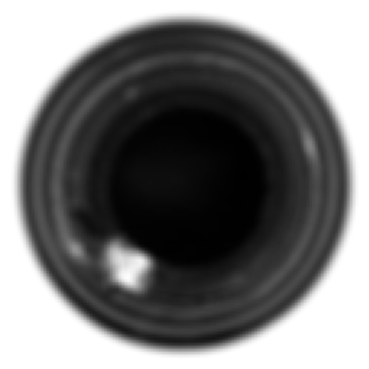}&
        % \hspace{-0.25cm}\includegraphics[width=0.11\textwidth]{figures/appendix_examples/bottle/contamination/6O.png}&
        \hspace{-0.25cm}\includegraphics[width=0.11\textwidth]{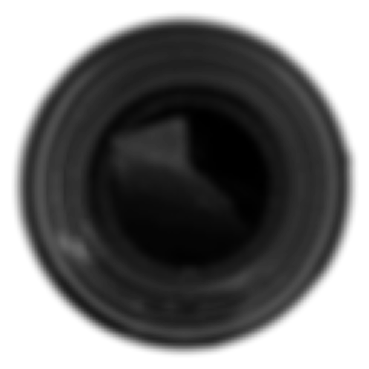}&
        \hspace{-0.25cm}\includegraphics[width=0.11\textwidth]{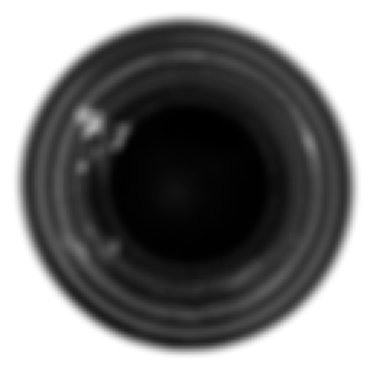}&
        \hspace{-0.25cm}\includegraphics[width=0.11\textwidth]{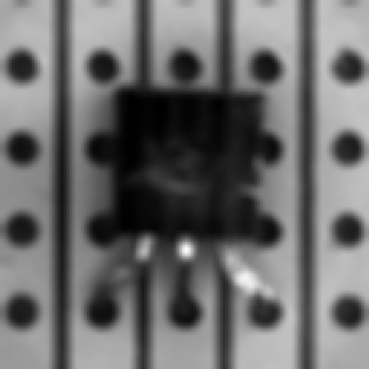}&
        \hspace{-0.25cm}\includegraphics[width=0.11\textwidth]{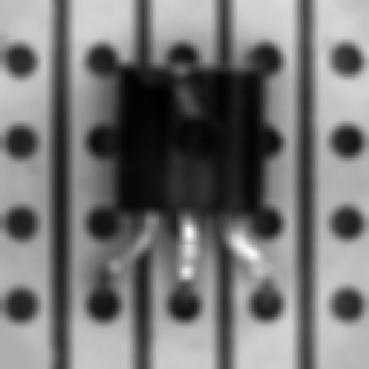}
        \\
        \multicolumn{8}{c}{(a) Original} 
        % \\
        % \includegraphics[width=0.11\textwidth]{figures/appendix_examples/hazelnut/crack/7R.png}&
        % \hspace{-0.25cm}\includegraphics[width=0.11\textwidth]{figures/appendix_examples/hazelnut/cut/13R.png}&
        % \hspace{-0.25cm}\includegraphics[width=0.11\textwidth]{figures/appendix_examples/hazelnut/print/11R.png}&
        % \hspace{-0.25cm}\includegraphics[width=0.11\textwidth]{figures/appendix_examples/bottle/contamination/6R.png}&
        % \hspace{-0.25cm}\includegraphics[width=0.11\textwidth]{figures/appendix_examples/bottle/contamination/0R.png}&
        % \hspace{-0.25cm}\includegraphics[width=0.11\textwidth]{figures/appendix_examples/bottle/brokenSmall/1R.png}&
        % \hspace{-0.25cm}\includegraphics[width=0.11\textwidth]{figures/appendix_examples/transistor/damagedCase/4R.png}&
        % \hspace{-0.25cm}\includegraphics[width=0.11\textwidth]{figures/appendix_examples/transistor/damagedCase/6R.png}
        % \\
        % \multicolumn{8}{c}{(a) Reconstruction}         
        \\
        \includegraphics[width=0.11\textwidth]{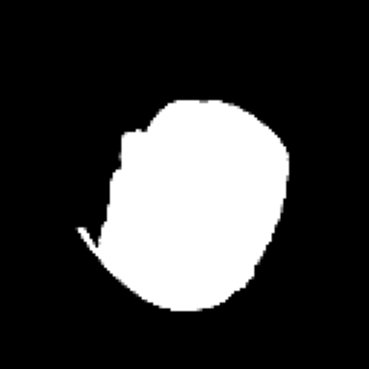}&
        \hspace{-0.25cm}\includegraphics[width=0.11\textwidth]{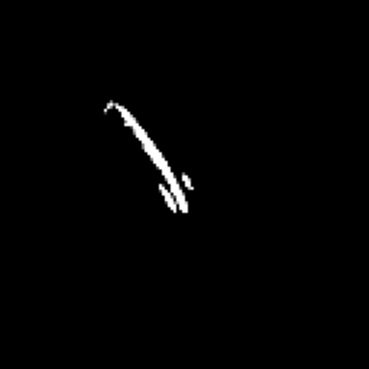}&
        \hspace{-0.25cm}\includegraphics[width=0.11\textwidth]{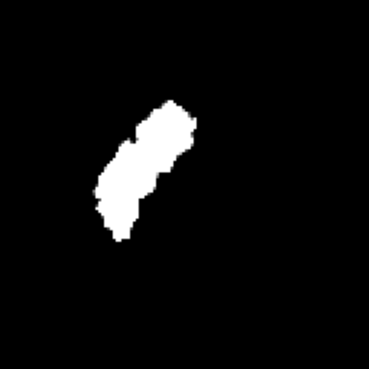}&
        \hspace{-0.25cm}\includegraphics[width=0.11\textwidth]{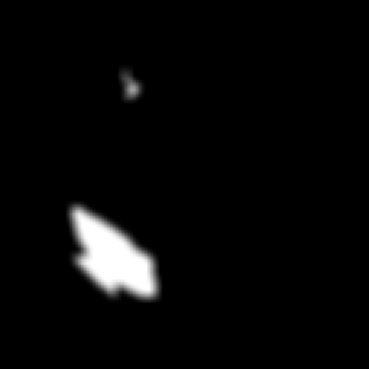}&
%        \hspace{-0.25cm}\includegraphics[width=0.11\textwidth]{figures/appendix_examples/bottle/contamination/6GT.png}&       
        \hspace{-0.25cm}\includegraphics[width=0.11\textwidth]{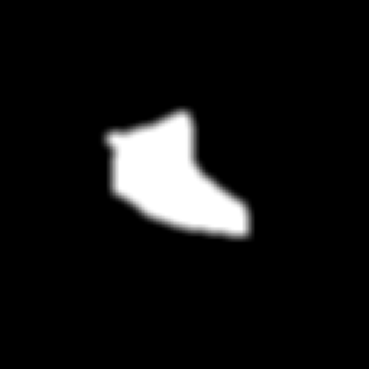}&
        \hspace{-0.25cm}\includegraphics[width=0.11\textwidth]{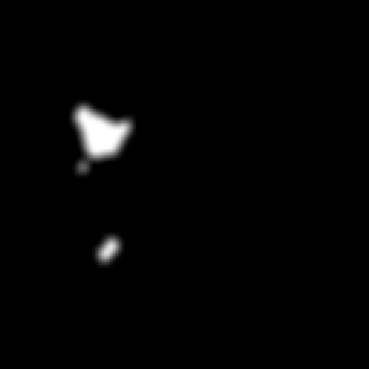}&
        \hspace{-0.25cm}\includegraphics[width=0.11\textwidth]{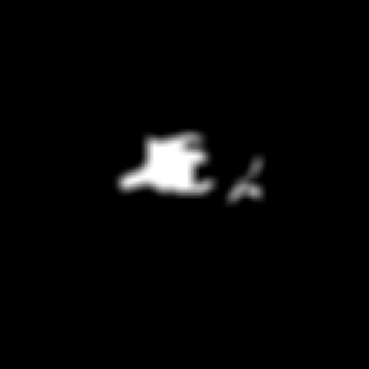}&
        \hspace{-0.25cm}\includegraphics[width=0.11\textwidth]{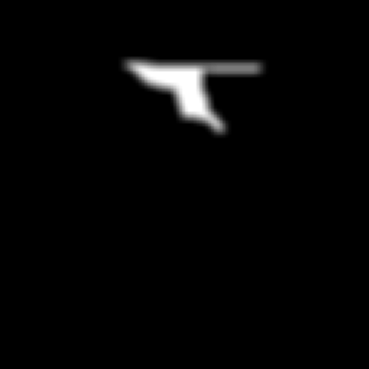}
        \\
        \multicolumn{8}{c}{(b) Ground-truth} 
        \\        
        \includegraphics[width=0.11\textwidth]{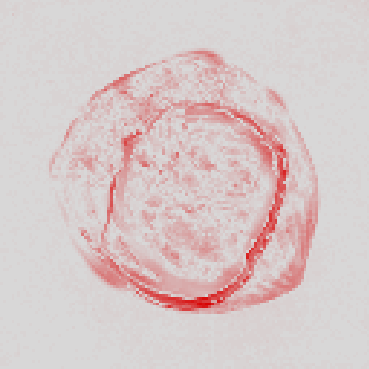}&
        \hspace{-0.25cm}\includegraphics[width=0.11\textwidth]{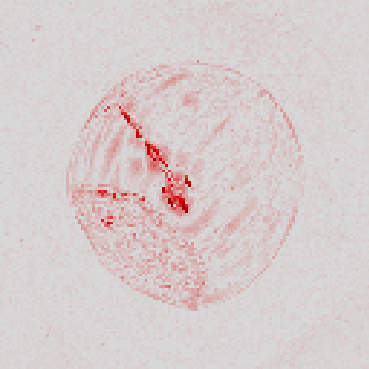}&
        \hspace{-0.25cm}\includegraphics[width=0.11\textwidth]{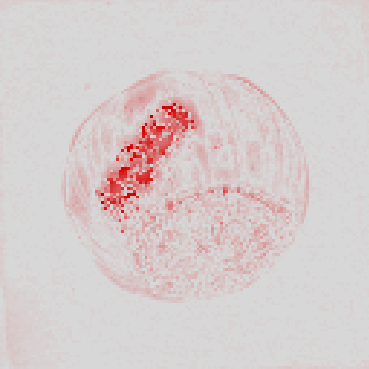}&
        \hspace{-0.25cm}\includegraphics[width=0.11\textwidth]{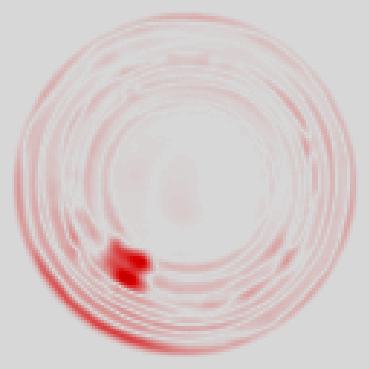}&
        %\hspace{-0.25cm}\includegraphics[width=0.11\textwidth]{figures/appendix_examples/bottle/contamination/6_heatmap L1.png}&
        %
        \hspace{-0.25cm}\includegraphics[width=0.11\textwidth]{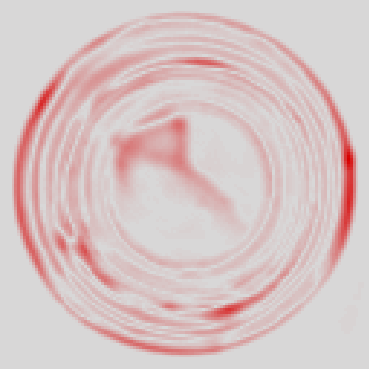}&
        \hspace{-0.25cm}\includegraphics[width=0.11\textwidth]{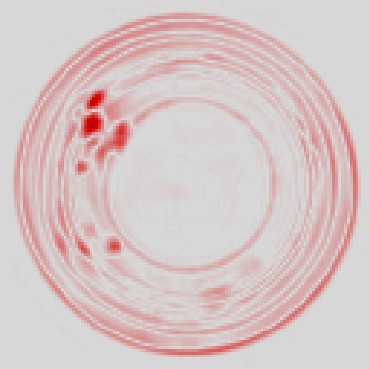}&
        \hspace{-0.25cm}\includegraphics[width=0.11\textwidth]{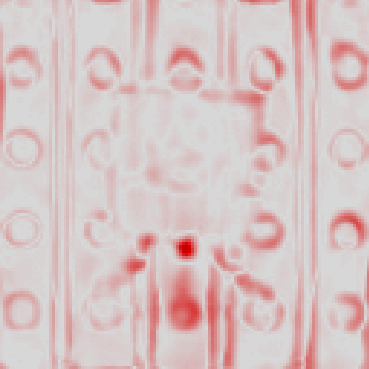}&
        \hspace{-0.25cm}\includegraphics[width=0.11\textwidth]{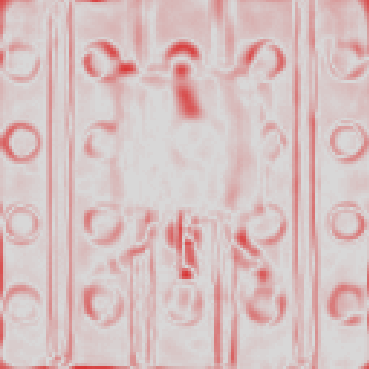}
        \\
        \multicolumn{8}{c}{(c) Residual-$L_1$} 
        \\
        \includegraphics[width=0.11\textwidth]{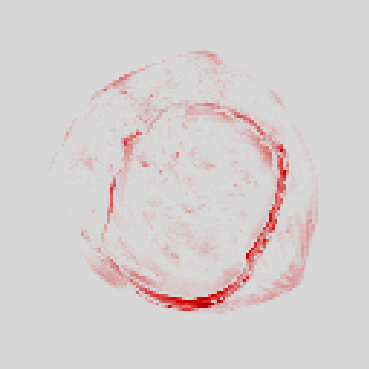}&
        \hspace{-0.25cm}\includegraphics[width=0.11\textwidth]{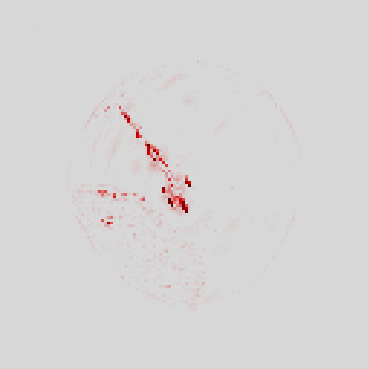}&
        \hspace{-0.25cm}\includegraphics[width=0.11\textwidth]{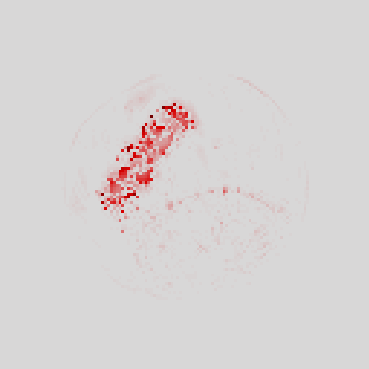}&
        \hspace{-0.25cm}\includegraphics[width=0.11\textwidth]{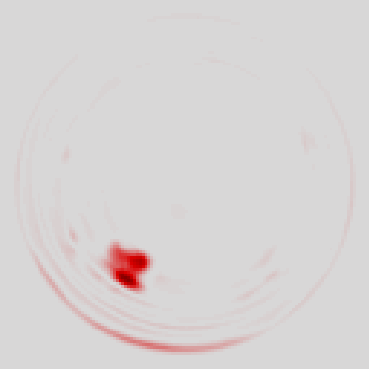}&        
        \hspace{-0.25cm}\includegraphics[width=0.11\textwidth]{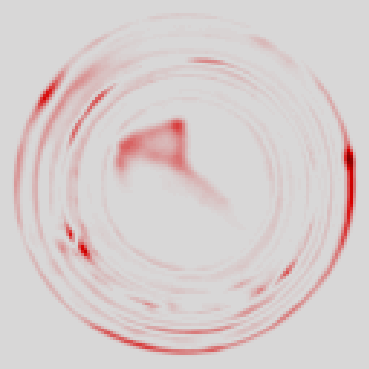}&
        \hspace{-0.25cm}\includegraphics[width=0.11\textwidth]{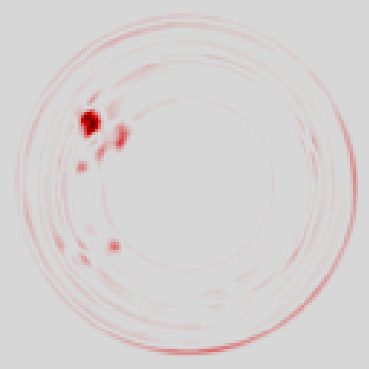}&
        \hspace{-0.25cm}\includegraphics[width=0.11\textwidth]{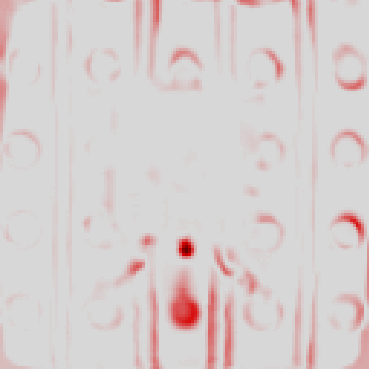}&
        \hspace{-0.25cm}\includegraphics[width=0.11\textwidth]{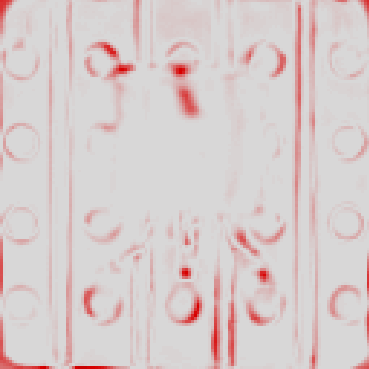}
        \\
        \multicolumn{8}{c}{(d) Residual-$L_2$} 
        \\
        \includegraphics[width=0.11\textwidth]{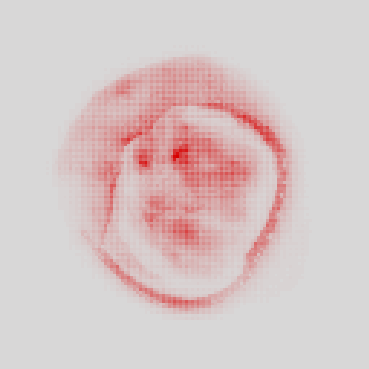}&
        \hspace{-0.25cm}\includegraphics[width=0.11\textwidth]{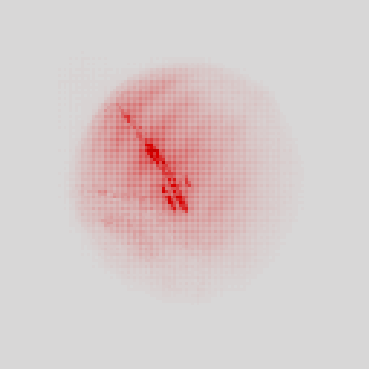}&
        \hspace{-0.25cm}\includegraphics[width=0.11\textwidth]{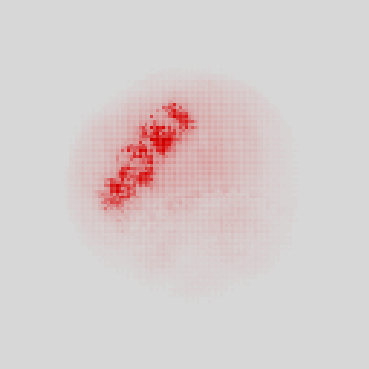}&
        \hspace{-0.25cm}\includegraphics[width=0.11\textwidth]{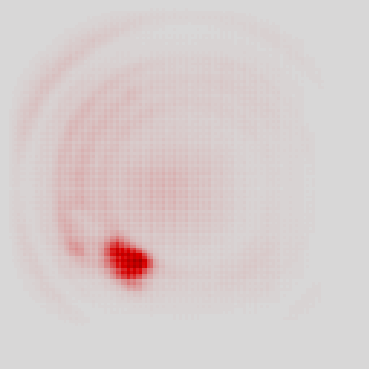}&
        \hspace{-0.25cm}\includegraphics[width=0.11\textwidth]{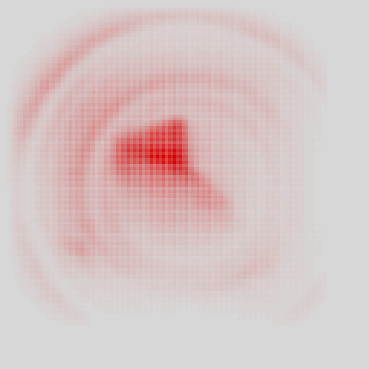}&
        \hspace{-0.25cm}\includegraphics[width=0.11\textwidth]{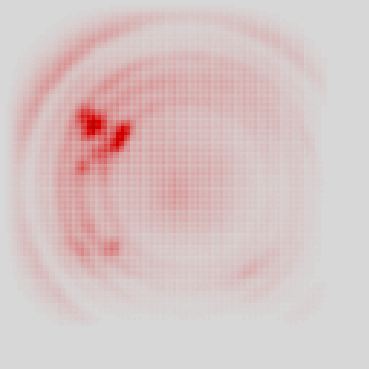}&
        \hspace{-0.25cm}\includegraphics[width=0.11\textwidth]{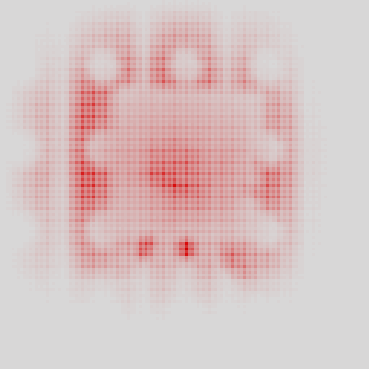}&
        \hspace{-0.25cm}\includegraphics[width=0.11\textwidth]{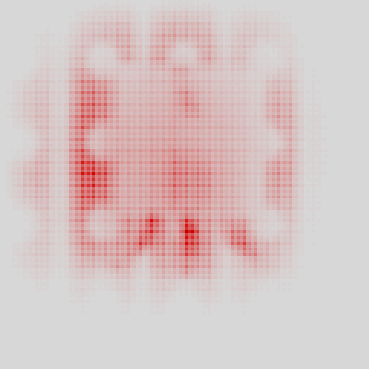}
        \\
        \multicolumn{8}{c}{(e) Ours-$L_1$}         
        \\
        \includegraphics[width=0.11\textwidth]{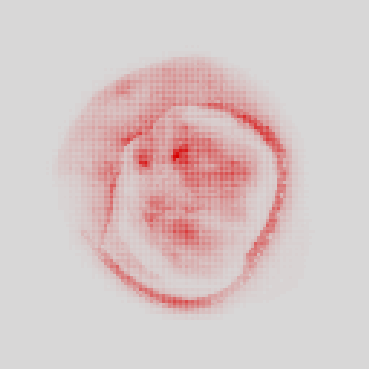}&
        \hspace{-0.25cm}\includegraphics[width=0.11\textwidth]{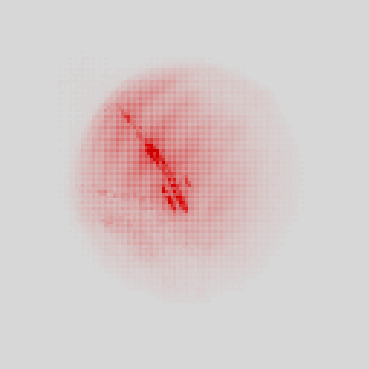}&
        \hspace{-0.25cm}\includegraphics[width=0.11\textwidth]{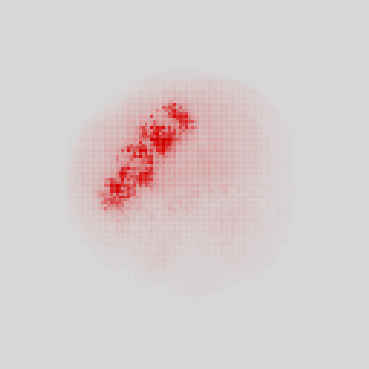}&
        \hspace{-0.25cm}\includegraphics[width=0.11\textwidth]{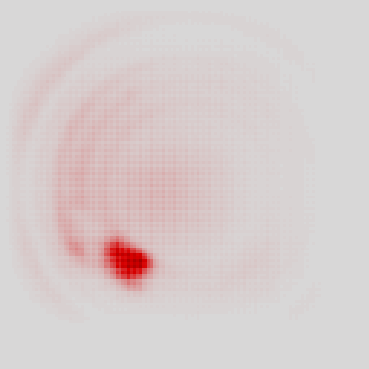}&
        \hspace{-0.25cm}\includegraphics[width=0.11\textwidth]{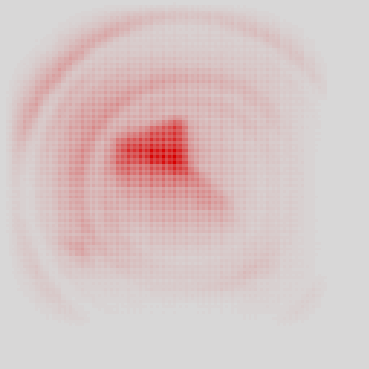}&
        \hspace{-0.25cm}\includegraphics[width=0.11\textwidth]{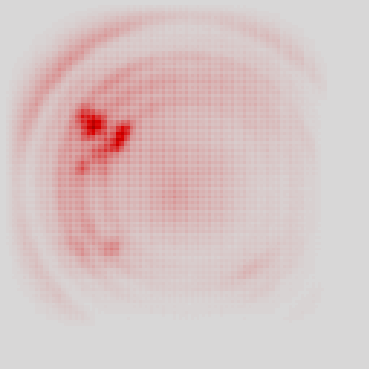}&
        \hspace{-0.25cm}\includegraphics[width=0.11\textwidth]{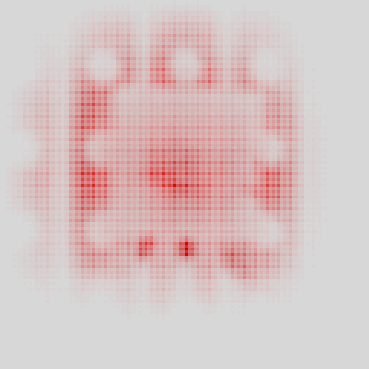}&
        \hspace{-0.25cm}\includegraphics[width=0.11\textwidth]{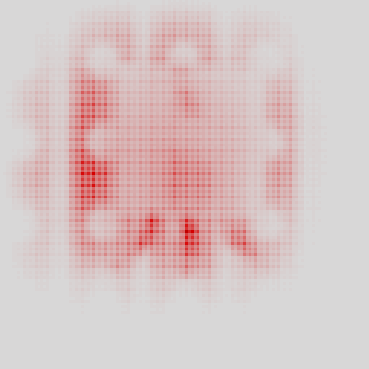}
        \\
        \multicolumn{8}{c}{(f) Ours-$L_2$} 
        \\
    \end{tabular}
    \caption{Explanations produced for images of damaged objects from the MVTec dataset~\cite{anomaly_image_dataset}. The figure illustrates the (a) original image; (b) ground truth damaged area; Autoencoder's explanations produced by the (c) baseline Residual-$L_1$ and (d) Residual-$L_2$ explainability methods and (e,f) results of the proposed LRP-based approach. For our approach, we show explanations generated by propagation of (e) $L_1$ and (f) $L_2$ reconstruction losses.} 
    \label{fig:appendix_result}
\end{figure*}

In this section we present additional explanations of anomalous samples from the MVTec dataset presented in Sec. \ref{sec:image_anomaly_detection}. Figure \ref{fig:appendix_result} follows the same schema as Figure \ref{fig:visual_result}.

\end{document}